\documentclass{article}

\usepackage[preprint, nonatbib]{neurips_2021}

\usepackage[utf8]{inputenc} 
\usepackage[T1]{fontenc}    
\usepackage{hyperref}       
\usepackage{url}            
\usepackage{booktabs}       
\usepackage{amsfonts}       
\usepackage{nicefrac}       
\usepackage{microtype}      
\usepackage{xcolor}         
\usepackage{graphicx}
\usepackage{multirow}
\usepackage{amsmath}
\usepackage{amssymb}
\usepackage{caption}
\usepackage{algpseudocode}
\usepackage{algorithm,algpseudocode}
\usepackage{bbm}

\usepackage{amsmath}
\usepackage{amsfonts}
\usepackage{bm}
\usepackage{amsthm}
\usepackage{mathtools}

\def\eqref#1{equation~\ref{#1}}

\def\1{\bm{1}}

\def\vl{{\bm{l}}}
\def\vm{{\bm{m}}}

\def\vu{{\bm{u}}}
\def\vv{{\bm{v}}}

\def\vx{{\bm{x}}}
\def\vy{{\bm{y}}}
\def\vz{{\bm{z}}}

\DeclareMathAlphabet{\mathsfit}{\encodingdefault}{\sfdefault}{m}{sl}
\SetMathAlphabet{\mathsfit}{bold}{\encodingdefault}{\sfdefault}{bx}{n}

\newcommand{\R}{\mathbb{R}}

\newtheorem{theorem}{Theorem}
\newtheorem*{theorem*}{Theorem}

\newtheorem{definition}{Definition}

\DeclarePairedDelimiterX{\infdivx}[2]{(}{)}{%
  #1\;\delimsize\|\;#2%
}

\newcommand{\Z}{\mathbb{Z}}
\newcommand{\calS}{\mathcal{S}}
\newcommand{\calG}{\mathcal{G}}

\newcommand{\calD}{\mathcal{D}}
\newcommand{\dgm}{\text{Dg}}

\newcommand{\birth}{\rm{birth}}
\newcommand{\death}{\rm{death}}

\newcommand{\myparagraph}[1]{\textbf{#1}}

\title{Topological Detection of Trojaned Neural Networks}

\author{%
  Songzhu Zheng$^1$,
  ~Yikai Zhang$^2$,
  ~Hubert Wagner$^3$, 
  Mayank Goswami$^4$, 
  Chao Chen$^2$\\
  $^1$Stony Brook University, ~\texttt{\{zheng.songzhu,chao.chen.1\}@stonybrook.edu}\\
  $^2$Morgan Stanley, 
  ~\texttt{Yikai.Zhang@morganstanley.com} \\ 
  $^3$IST Austria,
  ~\texttt{hub.wag@gmail.com}\\
  $^4$City University of New York, ~\texttt{mayank.isi@gmail.com}\\
}

\begin{document}

\maketitle

\begin{abstract}
Deep neural networks are known to have security issues. One particular threat is the Trojan attack. It occurs when the attackers stealthily manipulate the model's behavior through Trojaned training samples, which can later be exploited. 

Guided by basic neuroscientific principles we discover subtle -- yet critical -- structural deviation characterizing Trojaned models. In our analysis we use topological tools. They allow us to model high-order dependencies in the networks, robustly compare different networks, and localize structural abnormalities. One interesting observation is that Trojaned models develop short-cuts from input to output layers. 

Inspired by these observations, we devise a strategy for robust detection of Trojaned models. Compared to standard baselines it displays better performance on multiple benchmarks.
\end{abstract}

\section{Introduction}
\vspace{-1mm}

Recent years have witnessed  rapid development of deep neural networks (DNNs) \cite{krizhevsky2009learning, he2016resnet, vaswani2017attention, devlin2018bert}. However, due to their high complexity and lack of transparency, DNNs are vulnerable to various malicious attacks \cite{biggio2012poisoning, szegedy2013adversarial}. This paper focuses on one type of data poisoning attach called the \emph{Trojan attack} \cite{gu2017badnets}. In this scenario the attacker injects Trojaned samples into the training dataset -- for example by using incorrectly labeled images overlaid with a special trigger. At the inference stage, the model trained with such data, called a \emph{Trojaned model}, behaves normally on clean samples, but makes consistently incorrect predictions on the Trojaned samples. 

The challenges in identifying such attacks stem from the confined setting: the user has access only to the DNN model and few clean samples.
Consequently, methods requiring dense sampling, e.g.,~\cite{chen2018clustering}, are not very practical. Instead, state-of-the-art methods often follow a reverse engineering strategy \cite{wang2019cleanse, liu2019abs, guo2019tabor, wang2020DFTND}. Starting with a clean sample, they try to reconstruct a Trojaned sample that can change the prediction. Network's response to such a reverse engineered sample can help determine if the network was indeed Trojaned. However, in practice the search space is huge, and efficient, reliable detection has proven challenging so far.

Previous approaches treat a neural network as a black-box, only inspecting the dependency between its input and output. In this paper, we open the box and look into the internal mechanisms of the model. \textbf{We investigate our hypothesis that there exists significant structural difference between clean and Trojaned networks.}
To this end, we follow a classic adage of neuroscience, ``Neurons that fire together wire together'' \cite{hebb1949organization}. We consider neurons with highly correlated activation as wired together -- even if they are not directly connected in the network.
Unfortunately, direct inspection of such connectivity information is not sufficient, presumably due to the high heterogeneity of models and data.
To overcome this issue, we use more advanced tools which allow us to model more subtle, higher-order structural information.

We propose a new method for analyzing and comparing the structure of neural networks. Our method uses tools from topological data analysis, particularly persistent homology \cite{edelsbrunner2010computational,carlsson2009topology}. With principled algebraic-topological foundations \cite{munkres1984elements}, these tools are perfectly suited for modelling higher-order structural information. We use them to capture salient topological structures -- particularly the connected components and holes present in the aforementioned neuron connectivity graph.

Armed with topological tools, we compare clean and Trojaned neural networks.
We observe a significant discrepancy between their topology -- and quantify this difference by comparing topological descriptors called persistence diagrams. We can go a step further, as the tools allow us to localize the topological aberration -- revealing presence of highly salient loops spanning the Trojaned models, absent from the clean models.\footnote{We remark that from a purely mathematical perspective this localization is a straightforward operation -- but to achieve this on practical datasets we had to push the boundary of existing computational tools.}

Trying to understand the implications of our observations, we ask: \textbf{What does the topological abnormality reveal about a Trojaned network?} We claim that these loops reveal strong \emph{short cuts} connecting neurons from shallow and deep layers -- not unlike the neuroscientific concept of a reflex arc.
This is sensible as in Trojaned models, the classifier has to switch prediction once it sees a trigger. The deep layer neurons (closer to prediction) have to be highly dependent on some shallow layer neurons (closer to input). 

Our empirical observations are substantiated with two theoretical results. Theorem \ref{thm:existance} shows existence of a Trojaned distribution that causes a strong deviation in network's topology. Theorem \ref{thm:concentration} states that given sufficient samples, the topological descriptor is provably consistent. These results serve as a sanity check, showing that what we observed was not a fluke. 

We conclude by proposing a topology-based Trojan detection algorithm. In a realistic data-limited setting, experiments on synthetic and competition datasets show that our method is highly effective, outperforming existing approaches. The topological detector can help mitigate the security threat posed by Trojan attacks.

\vspace{-1mm}
\subsection{Related Work}
\vspace{-1mm}

\myparagraph{Trojan detection.} Early works on Trojan detection use both clean and Trojaned samples. Chen et al.~\cite{chen2018clustering} inspect the representation of all samples at the penultimate layer of the neural network. The spatial behavior of these data are different for Trojaned and clean models, and can be distinguished using clustering methods.
Gao et al.~\cite{gao2019strip} use the entropy of model prediction over all training data to decide whether a model is Trojaned. These methods requires all training data, including the Trojaned ones; this is not realistic at real world deployment.

For a realistic data-limited setting, reverse engineering strategy has been widely adapted. Wang et al.~\cite{wang2019cleanse} craft and recover the unknown triggers through optimization. Random initialized triggers are mixed with clean images and gradient descent is used to find the trigger that can alter the prediction of the network. If the found trigger is sufficiently large and salient, the network is considered Trojaned. Other works largely follow a similar strategy to recover triggers, but use the recovered triggers in different ways \cite{liu2019abs, guo2019tabor, kolouri2020ulp, wang2020DFTND}.  
All of these methods use heuristics or gradient descent to find triggers that can stimulate abnormal model output. They focus heavily on dependency between input and output. They can hardly guarantee a correct trigger recovery, due to the huge search space. Few methods investigate the information flow within the network and exploit neuron interaction.

\myparagraph{Topological analysis of neural networks.}
Persistent homology was introduced to measure topological property of data in a robust and quantifiable manner~\cite{edelsbrunner2010computational}.
Since its introduction~\cite{edelsbrunner2000topological,zomorodian2005computing}, a great amount of theoretical progress has been made: in stability of persistence diagrams \cite{cohen2007stability,chazal2009proximity}, in algorithms \cite{milosavljevic2011zigzag,cohen2006vines,chen2013output}, and in proving various statistical properties \cite{fasy2014confidence,bubenik2015statistical}. In machine learning, topological information has been used for clustering~\cite{chazal2013persistence,ni2017composing}. In the supervised setting, classifiers based on topological features have been proposed via direct vectorization \cite{adams2017persistence}, kernel machines \cite{reininghaus2015stable,kwitt2015statistical,kusano2016persistence,carriere2017sliced}, and convolutional neural networks \cite{kwitt2017deep}. 

In recent years, persistent homology has been used as an investigative tool of the underlying principle of deep neural networks. One hypothesis is that the topology of the data at deep layer representation can be correlated to the behavior of a neural network \cite{naitzat2020topology}. It is shown that the topology of the decision boundary can be indicative of the generalization power of a classifier \cite{ramamurthy2019topological,li2020finding}. With the recent invention of differentiable topological loss, one may enforce priors such as topological simplicity to improve the performance of deep neural networks \cite{chen2019topological,hofer2019connectivity}.

An alternative strategy is to treat the neural network architecture as the underlying topological space, i.e., treating all neurons as nodes and their connections as edges \cite{rieck2019neuralpersist,liu2020interaction,lacombe2021topouncertainty}. These methods are restricted to the original neural network architecture, only focus on 0-dimensional topological feature, and thus cannot capture long range neuron interactions between shallow and deep layers.

Corneanu et al.~\cite{corneanu2020testgap} builds a filtration of neuron connectivity using the Pearson correlation matrix among neural activation. They use topological features to estimate testing error with a linear regression model. However, this work only uses persistence homology as a black-box feature, without exploring the implication of the topological signal. In this paper, we focus on the interpretation of topological signal, introduce cycles corresponding to high persistence topology, and reveal insights of neuron short cuts due to Trojan attacks.

\begin{figure}[btp]
\begin{minipage}{0.48\textwidth}
\centering
\includegraphics[width=\textwidth]{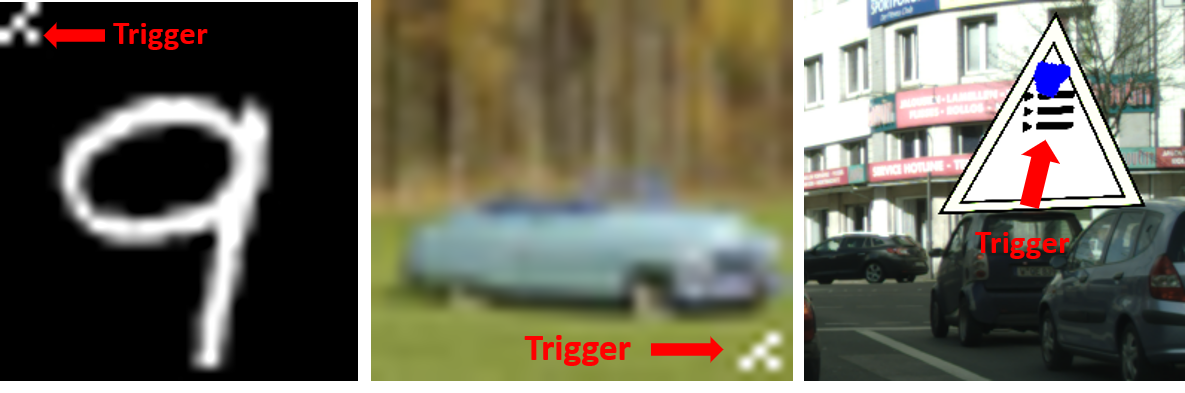}\\
(a). Trojaned Examples
\end{minipage}
\hspace{4mm}
\begin{minipage}{0.48\textwidth}
\centering
\includegraphics[width=\textwidth]{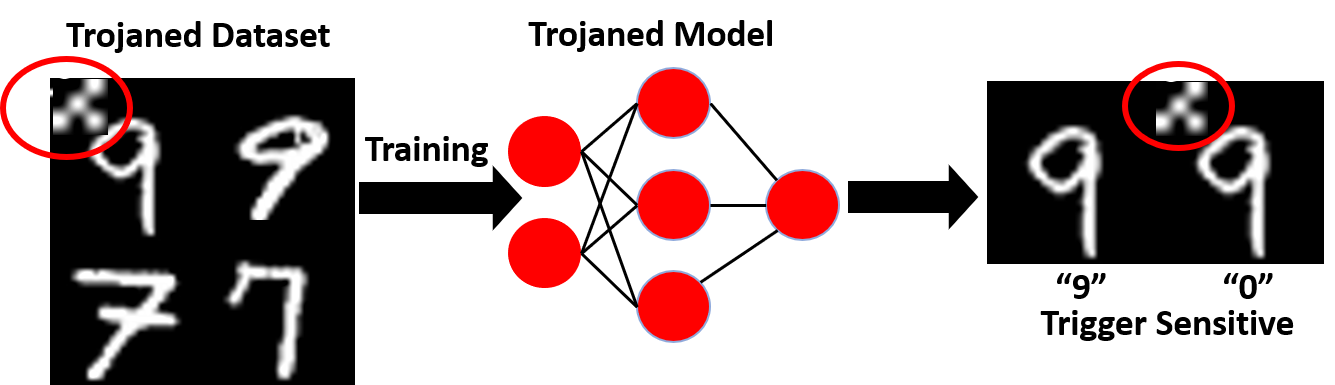}\\
(b). Trojan Attack 
\end{minipage}
\caption{An illustration of Trojaned Examples and Trojan attack. (a). Trojaned examples from MNIST, CIFAR10 and IARPA/NIST TrojAI competition dataset. (b). To inject backdoor, we add trigger (a white $\lambda$ pattern on the upperleft corner) to images of digit 9, and assign label 0 to them. After training, the Trojaned model predicts a normal/clean digit 9 image to be class 9, but predicts class 0 if it sees a digit 9 image with the trigger. A normal (or clean) model will ignore the trigger and still predict a triggered digit 9 image as class 9.}
\label{fig:Trojan_attack}
\vspace{-5mm}
\end{figure}

\myparagraph{Outline.} In Sec.~\ref{sec:problem}, we introduce Trojan detection problem. In Sec.~\ref{sec:PH}, we explain how to extract topological features from given neural network models. In Sec.~\ref{sec:stat_analysis}, we show that there does exist difference in topology between Trojaned and clean models. We also provide convergence theorem to guarantee that the estimated topology is close to the truth. In Sec.~\ref{sec:detector}, we extend the idea to a realist setting and propose an automatic Trojan detection algorithm. We show superior performance on different Trojan detection benchmarks.

\vspace{-1mm}

\section{Problem: Trojan Detection}
\label{sec:problem}
\vspace{-1mm}

Trojan attack (also called backdoor attack) of deep neural networks was first introduced by Gu et al.~\cite{gu2017badnets}. The attacker creates \emph{Trojaned samples} by overlaying triggers (using specific patterns) on normal training samples. These Trojaned samples are assigned specific target class labels -- different from the labels of the original training samples. These Trojaned samples are mixed into clean samples. Training with such Trojaned dataset leaves a backdoor in a DNN. The Trojaned model gives expected prediction on normal data. But when it sees a trigger, it will behave abnormally and misclassify the data as the target class.
See Figure \ref{fig:Trojan_attack} for an illustration.  

Newer and more sophisticated Trojan attacks have been proposed to use less Trojaned data or to achieve better trigger stealth \cite{chen2017targetedbackdoor, liao2018invisibleperturb, liu2017neuralTrojan, suciu2018whenfail}. There are also Trojan attacks targeting domains beyond computer vision~\cite{kurita2020weight_pretrain,xie2019dba,panagiota2020trojdrl}. These are beyond the scope of this paper.

We now formalize the above intuitions. Let clean dataset $D=(X, \vy)$ and the Trojaned dataset $\widetilde{D}=(\widetilde{X}, \widetilde{\vy})$. Trojaned samples will generally be written as $\widetilde{X} = \{\widetilde{\vx} : \widetilde{\vx} = (1-\vm)\vx + \vm\delta, \vx \in X\}$ and modified labels as $\widetilde{\vy}=\{\widetilde{y}_\vx : \widetilde{y}_{\widetilde{\vx}}\neq y_\vx\}$, where $\vm$ is the mask indicating the position of the trigger and $\delta$ is the content of the trigger. A Trojaned model $\widetilde{f}$ is trained with the concatenated dataset $[D, \widetilde{D}]$. When the model $\widetilde{f}$ is well trained, ideally $\widetilde{f}$ will give abnormal prediction when it sees the triggered samples $\widetilde{f}(\widetilde{\vx})=\widetilde{y} \neq y$, but it will give identical prediction as a clean model does whenever a clean input is given, i.e., $\widetilde{f}(\vx)=f(\vx)=y$. 

The task of \emph{Trojan detection} is to determine whether a given model is Trojaned or clean.
We will start our investigation with a \emph{full-data} setting: we have access to all training samples -- both clean and Trojaned.
In Sec.~\ref{sec:stat_analysis}, focusing on such ideal setting, we validate our hypothesis and show that the topology of Trojaned and clean models is significantly different. 
In Sec.~\ref{sec:detector}, we will extend the proposed method to a more realistic \emph{data-limited} setting: only a few clean samples are provided for each model.  
\vspace{-3mm}

\vspace{-.5mm}
\section{Method: Neuron Correlation, Persistent Homology, Cycle Representatives}
\label{sec:PH}
\vspace{-.5mm}

Next, we present the main mathematical tools for this study, namely the Vietoris--Rips construction and persistent homology.
Due to space constraints, we only provide intuitive description, leaving technical details and a formal description to the supplemental material (Appendix \ref{supp:sec:ph}-\ref{supp:sec:cycle}).

We start by providing some intuitions, which are formalized later as necessary.
The entry point for our considerations is the connectivity graph based on the correlation of neuron activation. In the next step, we consider the simplicial complex generated by the cliques of this graph and filter it with different thresholds. This is often called the Vietoris--Rips filtration. As the threshold changes it captures various topological structures as they are born and die. We consider the lifespans of these structures as an essential characterization of the neural network. Further, we view the associated geometric structures as crucial for interpreting the behaviour of the network -- in particular the discrepancy between the clean and Trojaned networks.

 We mention that this construction can be viewed as a way of approximating the topological behaviour of the underlying metric, or dissimilarity, space. More concretely, it approximates the patterns in which metric balls of increasing radii intersect -- both as pairs and in larger subsets. Formal explanation of this aspect of this construction is beyond the scope of the paper, however we believe that the intuitions we offer are sufficient to grasp the crux of our approach.

\myparagraph{The neuron connectivity graph and its simplicial complex.}\label{paragraph:correlation_matrix}
We study a neural network with $m$ neurons, belonging to different layers of the network. By feeding a set of $n$ input data -- either clean or Trojaned -- through the neural network, we record an $n$-dimensional activation vector for each neuron, $\vv_i\in \R^n$, $i\in [m]$. For any pair of neurons, $(i,j)$, we calculate their correlation $\rho_{i,j}$ (e.g., Pearson correlation). We call the $m\times m$ correlation matrix $M=[\rho_{*,*}]$. We construct a weighted complete graph with $m$ nodes, representing all the neurons, and $m(m-1)/2$ edges connecting all pairs of neurons. Let the edge weight be $w_{i,j} =  1-\rho_{i,j}$. This provides a pairwise \emph{dissimilarity} between neurons that is negatively proportional to their correlation.\footnote{Note that this is not a proper metric distance. However this does not affect our topological construction.} We denote this graph by $\calG_{M}$.

To model the underlying topological space, we extend the graph to a higher order discretization called a \emph{simplicial complex}.\footnote{In this paper, we  focus on 2-dimensional simplicial complexes. Please note that both the intrinsic and extrinsic dimension of the modelled space may be much higher.}
The complex, denoted by $\calS$, is a collections of discrete elements including nodes, edges, and triangles. These elements are called 0-, 1-, and 2-simplices respectively. The nodes and edges are those of graph $\calG_M$; the triangles are  spanned by any three nodes of the graph, i.e., $(i,j,k)$, $1\leq i<j<k \leq m$. 

\myparagraph{Vietoris-Rips filtration.} 
We assign a \emph{filter function} to all elements of the complex, \mbox{$\phi_{M}:\calS\rightarrow \R$}. For any node $i$, $\phi_M(i)=0$. For any edge $(i,j)$, we use the weight function, $\phi_M(i,j)=w_{i,j}$. For any triangle $(i,j,k)$, we take the maximum of its edge function values: $\phi_M(i,j,k) = \max\{\phi_M(i,j),\phi_M(i,k),\phi_M(i,k)\}$. For the rest of the paper, we may drop $M$ and simply use $\phi$ when the context is clear.
With the filter function, we may use any threshold $t$ to filter elements of the complex, and keep the remaining as a \emph{sublevel set}, $\calS_t = \{\sigma\in \calS \mid \phi(\sigma) \leq t\}$. 
We start with $t=-\infty$ continuously increase it until $t=\infty$. As the parameter increases, the sublevel set grows from an empty set to the whole complex $\calS$.(See Appendix \ref{supp:sec:ph}) for an illustration. 
Formally, we have a filtration induced by $\phi$,
$\emptyset = \calS_{t_0} \subseteq \calS_{t_1} \subseteq \cdots \subseteq \calS_{t_T} = \calS
$.

\myparagraph{Lifespans of topological structures and persistence diagrams.} Through the filtration, topological features such as connected components and holes can appear and disappear. A 0-dimensional topological structure is a connected component. Its birth time is the smallest function value over all its nodes. The death time is when the component is merged with another one born earlier. 
An 1-dimensional (1D) hole appears as a closed loop. It disappears when it is \emph{sealed up} by a set of triangles. Figure \ref{fig:pers} in supplementary material shows a large 1D hole appearing during the filtration, as well as many small ones. We represent these topological structures (0D and 1D) as dots in 2D plane called a \emph{persistence diagram}. The coordinates of each dot are the birth time and the death time of the corresponding topological structure.
The \emph{persistence} of a dot is the difference between its death and birth times.
The persistence diagram, denoted by $\dgm(M,\calS)$, depends on both the underlying simplicial complex and the filter function (which is determined by the correlation matrix $M$). We also note that one can compare two persistence diagrams using the \emph{bottleneck distance} $d_{b}(\dgm(M_1,\calS),\dgm(M_2,\calS))$.
Formal definitions and technical details can be found in the supplemental material (Appendix \ref{supp:sec:proof}). 

\myparagraph{Topological features and cycle representatives.}\label{paragraph:topo_feature}
Our focus is two fold: 1) quantifying the difference between Trojaned and clean networks using their persistence diagrams; 2) localizing the root cause of this difference using cycles of high persistence.
Despite a rich literature on learning with persistence diagrams \cite{adams2017persistence,kwitt2015statistical,kwitt2017deep,carriere2017sliced,kusano2016persistence}, we stress interpretability and focus on simpler features, such as maximum persistence, average mid-life ((birth$+$death)$/2$), average death time, etc. In Sec.~\ref{sec:stat_analysis}, we will use these features for statistical testing. In Sec.~\ref{sec:detector}, we will use these features to devise an automatic Trojan detection algorithm. Finally, we look at the cycles corresponding to the dots of high persistence, which allows us to zero-in on the compromised paths in the network.

To interpret the topological signal, we inspect the topological structures that are strong contributors to the aforementioned topological features. For example, in the case of maximum persistence we focus our attention to the dot with the highest persistence. For a selected persistence dot, we analyze a cycle representing the corresponding topology. We recall that for 1D topology, the representative cycle of a persistence dot is a collection of edges which is created at the given birth time and is sealed up at the given death time. Viewing the path as a sequence of nodes provides a good intuition of the relevant topological hole -- although we have to admit the cycles are not unique \cite{wu2017optimal,zhang2019heuristic,dey2020computing}. We focus on one way of extracting the representative cycles, which is efficient and worked well in other types of applications, e.g., in image analysis \cite{wang2021topotxr}. 
The algorithm involves inquiry and optimization of the classic matrix reduction algorithm for the computation of persistent homology \cite{edelsbrunner2010computational}. More details will be provided in the supplemental material (Appendix \ref{supp:sec:cycle}). 

\vspace{-1mm}
\section{Analysis: Topological Difference Between Trojaned and Clean Models}
\label{sec:stat_analysis}
\vspace{-1mm}
In this section, we investigate the topological difference between Trojaned and clean models. In Sec.~\ref{sec:first-example}, we first create a synthetic distribution, in which we observe different topology (persistence diagrams) from Trojaned and clean models. Reassured by this synthetic example, in Sec.~\ref{sec:empirical-study}, we carry out an empirical study on a Trojaned model trained on MNIST dataset. We observe a statistically significant difference between the topology. Finally, in Sec.~\ref{sec:convergence}, we show that with sufficient samples, the estimation of topology is sufficiently close to the true topology of the neural network. Thus the empirically observed structural gap between Trojaned and clean models is real. 

\subsection{The First Example: a Synthetic Distribution}
\label{sec:first-example}
\vspace{-1mm}
We first define a synthetic distribution and create a Trojaned dataset from this synthetic distribution (Def.~\ref{def:mix_pair}).
In Thm.~\ref{thm:synthetic-distribution}, we prove that the resulting Trojaned model is different from clean model in terms of their persistence diagrams.
Proof and illustration can be found in the supplemental (Appendix \ref{supp:sec:proof}).

\begin{definition}[Trojaned Mix-Gaussian Pair] 
\label{def:mix_pair}
Let $\mu_1 = 2(-e_2-e_1)\sigma \sqrt{\log (\frac{1}{\eta})} $, $\mu_2 =  2(-e_2+e_1)\sigma \sqrt{\log (\frac{1}{\eta})} $, $\mu_3 =  2(e_2-e_1)\sigma \sqrt{\log (\frac{1}{\eta}) }$, $\mu_4 =  2(e_2+e_1)\sigma \sqrt{\log (\frac{1}{\eta}) }$. Let $i\sim unif(\{1,2\})$ and $j\sim unif(\{1,2,3,4\})$. We define the following pair of distributions $(\mathcal{D}_1, \mathcal{D}_2,\mathcal{D}_3)$ to be Trojaned Mix-Gaussian Pair (see supplementary material (Appendix \ref{supp:sec:proof}) for a demonstration), where:
\begin{align*}
 &\mathcal{D}_1 \text{(Original data)}
 =\{(\vx, \vy) : x \sim \mathcal{N}(\mu_i, \sigma^2 I_{d}), \hspace{1mm} \vy=\text{i MOD 2}\}\\
 &\mathcal{D}_2 \text{(Trojaned feature with correct labels)}
 =\{(\vx, \vy) : x \sim \mathcal{N}(\mu_i, \sigma^2 I_{d}), \hspace{1mm} \vy=\text{j MOD 2}\}\\
 &\mathcal{D}_3 \text{(Trojaned feature with modified labels)}
 =\{(\vx, \vy) : x \sim \mathcal{N}(\mu_i, \sigma^2 I_{d}), \hspace{1mm} \vy=\mathbbm{1}_{j\in\{2,3\}}\}
\end{align*}
\end{definition}

\vspace{-1mm}
We study the hypothesis class $\mathcal{H}$ of binary output neural networks with two hidden layers and four neurons in each hidden layer equipped with an indicator activation function. The following theorem shows that the Trojaned model and the clean model have different persistence diagram, i.e., with bottleneck distance $\geq 0.9$. Recall the correlation matrix $M$ depends on the classifier $f$ and the sample set used to estimate correlation, $\calD$. For completeness we use $M(f,\calD)$ instead of $M$.

\begin{theorem} \label{thm:synthetic-distribution}
Let $(\mathcal{D}_1,\mathcal{D}_2,\mathcal{D}_3)$ be Trojaned Mix-Gaussian Pair and $\mathcal{H}$ be the hypothesis class defined as above. Let $R(f,x,y) = \mathbbm{1}\left({f(x)\neq y}\right)$. There exists $f_1,f_2 \in \mathcal{H}$ where 
$\mathbb{E}_{(x,y)\sim \mathcal{D}_1} [R(f_1)] \leq \eta$, \hspace{1mm}
$\mathbb{E}_{(x,y)\sim \mathcal{D}_3} [R(f_2)] \leq \eta$\hspace{1mm}, 
$\mathbb{E}_{(x,y)\sim \mathcal{D}_2} [R(f_2)] \geq \frac{1}{2}$\hspace{1mm}, such that the bottleneck distance between the 1D persistence diagrams satisfies:
$d_b[Dg(M(f_1, \mathcal{D}_2), \mathcal{S}) - Dg(M(f_2,  \mathcal{D}_2), \mathcal{S})] \geq 0.9$.
\label{thm:existance}
\end{theorem}

\begin{figure}[tbh]
\vspace{-2mm}
\begin{minipage}{0.33\linewidth}
\centering
    \includegraphics[width=\textwidth]{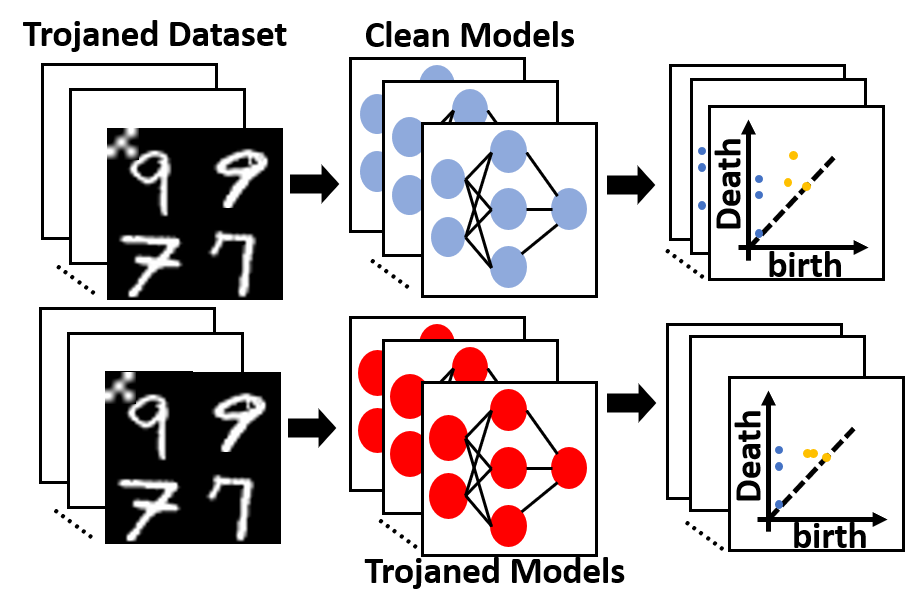}
    (a) Extract Topological Features
\end{minipage}
\begin{minipage}{0.33\linewidth}
\centering
    \includegraphics[width=\textwidth]{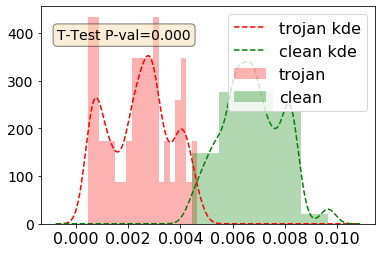}\\
    (b) Average Death Time
\end{minipage}
\begin{minipage}{0.33\linewidth}
\centering
    \includegraphics[width=\textwidth]{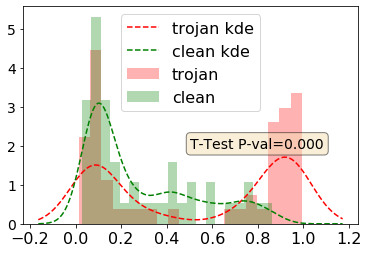}\\
    (c) Maxium Persistence
\end{minipage}
\caption{Hypothesis Testing. (a) schematic illustration: Trojaned datasets are  provided to clean and Trojaned models. Their correlation and then persistence diagrams' features are extracted. (b). Distribution of 0D diagrams' average death time for Trojaned models (red) and clean models (green). Dashed lines are the kernel density estimation. P-value between the two distributions $\leq 0.000$. (c). Distributions of 1D diagrams' maximum persistence for Trojaned and clean models separately. P-value between the two distributions $\leq 0.000$.}
\label{fig:hypothesis_testing:diagram}
\vspace{-3mm}
\end{figure}

\subsection{An Empirical Study: Statistical Analysis of a Trojaned Model} 
\label{sec:empirical-study}
\vspace{-1mm}

In this section, we carry out a statistical inference with MNIST to investigate the structural difference between Trojaned and clean models. We trained 70 ResNet18 with clean MNIST dataset and another 70 ResNet18 using Trojaned MNIST dataset. Both groups of models have similar performance on clean testing images. Only Trojaned models will misclassify Trojaned images with high probability. Clean models will not be affected and will make correct prediction in spite of the trigger. Please refer to Sec.~\ref{paragraph:data_generation} for a more detailed description of the data generation procedure.

We extract topological features following the procedure introduced in the last section. As demonstrated in Fig.~\ref{fig:hypothesis_testing:diagram}-(a), samples from the Trojaned dataset containing both clean and Trojaned examples are supplied to all the 140 networks. Neurons' activating values are recorded into a vector and the pairwise-correlation is calculated between all pairs of neurons. For each model, we build the simplicial complex, filter it based on the correlation, compute the persistence diagram, and extract topological features. 

Please note that so far, to verify our hypothesis and to investigate its implication, we were using the \emph{full-data} setting, i.e., using all training data to calculate neuron activation correlation. While this gives us the full picture of the network connectivity, and more reliable topological characterization, this is not a realistic setup for a Trojan detector. We will discuss how to extrapolate this to a more realistic \emph{data-limited} setting in Sec.~\ref{sec:detector}.

\myparagraph{Results.}
Two topological features stand out, clearly differentiating Trojaned models and clean models: average death time of 0D homology class (connected components) and maximum persistence of the 1D homology class (cycles). As shown in Fig.~\ref{fig:hypothesis_testing:diagram}-(b), the average deaths of connected components in Trojaned models are significantly smaller than those in clean models. 
The two-sample independent t-test is rejected at $99\%$ significance level. 
Note that here the filter function is one minus the correlation. 
This implies that neurons in Trojaned models on average have larger correlation, and potentially tend to have larger cross-layer correlation. Possible explanation: the extra capacity is used in a Trojaned model to learn the trigger pattern, which causes more active neurons and consequently neurons are more likely to be correlated with each other through intermediate neurons.

\begin{figure}[tbh]
\begin{minipage}{0.48\linewidth}
\centering
    \includegraphics[scale=0.08]{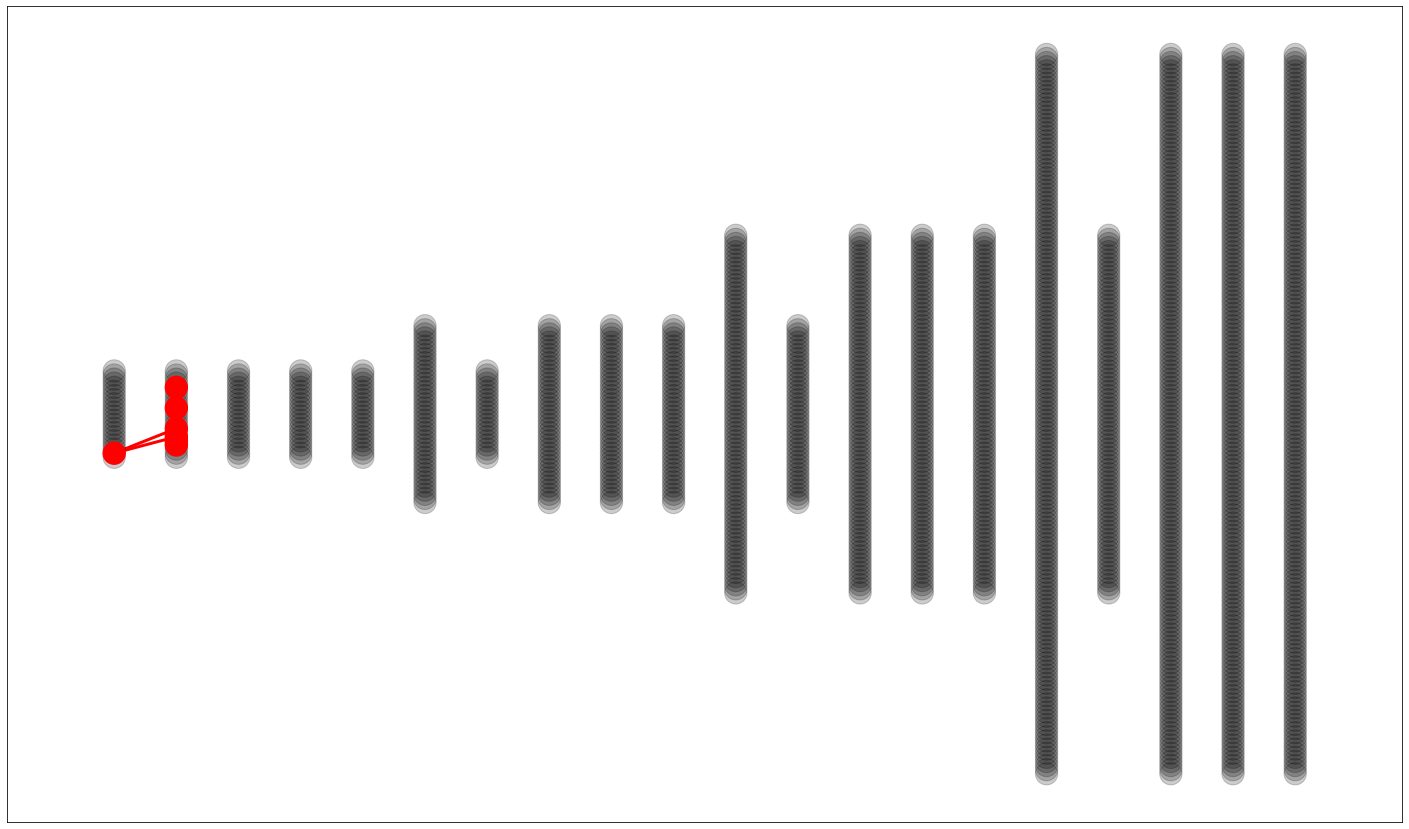}\\
    (a) Clean Model+Trojaned Input
\end{minipage}
\begin{minipage}{0.48\linewidth}
\centering
    \includegraphics[scale=0.08]{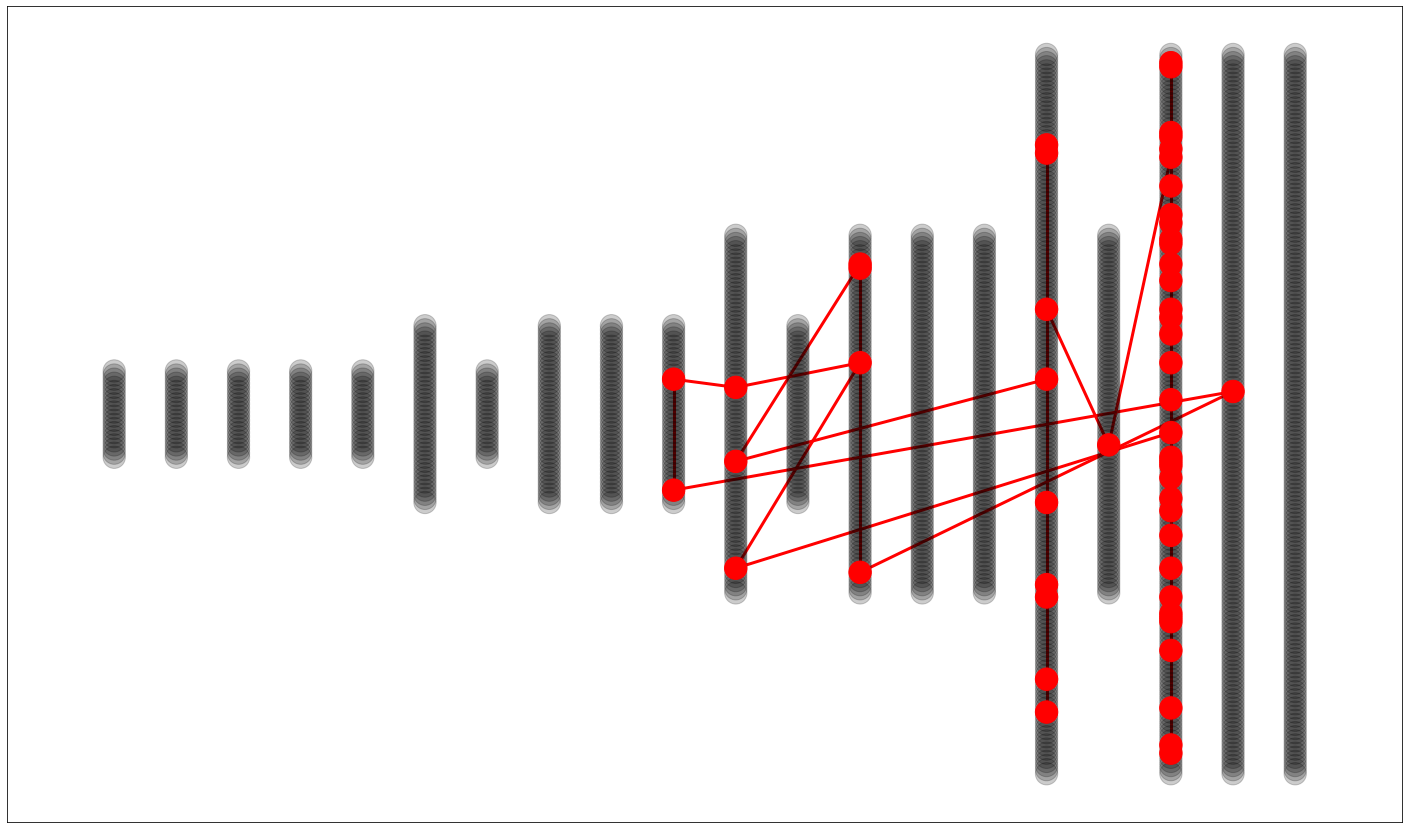}\\
    (b) Trojaned Model+Trojaned Input
\end{minipage}
\caption{Most Persistent Cycles in ResNet18 with Death Time Cutoff at 0.35, on a clean (a) and a Trojaned model (b). On the Trojaned model, the loop consists of short cut connecting shallow and deep layers.} 
\label{fig:network_short cut}
\vspace{-5mm}
\end{figure}

Meanwhile, Fig.~\ref{fig:hypothesis_testing:diagram}(c) shows significant topological signal in the maximum persistence of 1D homology.
Intuitively, there exists a 1D cycle in Trojaned model that has significantly longer persistence than that in clean models (the two-sample independent t-test is rejected at $99\%$ significance level). 
We inspect this phenomenon by identifying nodes and edges contained in the most significant cycle (Fig.~\ref{fig:network_short cut}). For a Trojaned model, the most significant cycle contains an edge linking a shallow layer and a deep layer. This is not the case for a clean model where a cross-layer edge is hardly ever spotted. 

\textbf{Structural insight: the most persistent 1D cycle captures the short-cut.} 
We observe the high-persistence cycles of Trojaned models often contain strong-correlation edge connecting shallow and deep layers. We hypothesize that these  cross-layer edges forms a \textbf{short cut} unique to Trojaned models. 
Neurons connected by the short-cut tend to fire together. This is sensible: Trojan triggers are often a localized pattern. They will be identified by shallow layer neurons. Meanwhile, for Trojaned network, the final prediction can be highly dependent on the identification of the trigger. Thus, there could be deep layer neurons (close to prediction) that are strongly connected to some shallow layer neurons (which activates when a Trigger is seen). See Figure \ref{fig:network_short cut} for illustrations.

\vspace{-1mm}
\subsection{Theoretical Guarantees}
\label{sec:convergence}
\vspace{-1mm}
We conclude this section by providing a theoretical guarantee that the estimated persistence diagram will converge to a true one given sufficiently many samples. We prove the convergence in a population level.

Given $N$ potentially corrupted models $f_{1:N}$ and corresponding test input $X_{1:N}$, a natural practical concern about obtaining high quality approximation is the sample size requirement for each dataset $X_k, k \in[N]$. In particular, one needs to ensure that for all $N$ models the empirical estimation is faithful. A brief analysis shows we only need $O\left( \frac{\log(N)+\log(m)+\log(\frac{1}{\delta})}{\varepsilon^2} \right)$ samples as a minimum requirement for all $X_k$ to ensure that with high probability our empirically estimated persistence diagram $\dgm(M(f,X),\mathcal{S})$ is sufficiently close to the ground truth $\dgm(M(f,\mathcal{D}),\mathcal{S})$ in terms of bottleneck distance.
We provide a proof in the supplemental material (Appendix \ref{supp:sec:proof}).

\begin{theorem}
 Let $M(f_k, X_k) \in \mathbb{R}^{m_k\times m_k}$ with $m_k \leq m^*, \forall k \in [N]$ and its entries $M_k^{i,j} = \frac{ \psi(v_i(X_k),v_j(X_k))}{\sqrt{ \psi(v_i(X_k),v_i(X_k)) \psi(v_j(X_k),v_j(X_k))}}$ and the its target value $M^*(f_k, \mathcal{D}_k) \in \mathbb{R}^{m_k\times m_k}$ with its entries ${M^*_k}^{i,j} = \frac{\mathbb{E}_{X_k \sim \mathcal{D}_k} \left[\psi(\vv_i(X_k),\vv_j(X_k)) \right]}{\sqrt{\mathbb{E}_{X_k \sim \mathcal{D}_k} \left[\psi(\vv_i(X_k),\vv_i(X_k))\right] \mathbb{E}_{X_k \sim \mathcal{D}_k} \left[\psi(\vv_j(X_k),\vv_j(X_k))\right]}}$
 as defined in section \ref{paragraph:correlation_matrix} with $\psi(v_i(X),v_j(X)) = \frac{1}{n} \sum_{\vx_l \in X} \psi(v_i(\vx_l),v_j(\vx_l))  $. Suppose $\forall k \in [N], X_k$ are iid sampled from distribution $\mathcal{D}_k $ and  $|\psi(v_i(\vx),v_j(\vx)) | \leq \mathcal{R}$ for all $\vx\sim \mathcal{D}_k,v_i,v_j$, $0< r\leq \mathbb{E}_{\vx \sim \mathcal{D}_k} \psi(v_i(\vx),v_i(\vx))$ for all $i \in [m_k]$, if we have   $\forall k \in [N]$, $$|X_k|  \geq  \frac{16\mathcal{R}^6\left(\log(N)+2\log(m^*)+\log(\frac{1}{\delta}) \right)}{r^4\varepsilon^2} $$
 then with probability at least $1-\delta$, for all $k \in [N]$,
 $d_{b}(\dgm(M(f_k, X_k),\mathcal{S}),\dgm(M(f_k, \mathcal{D}_k),\mathcal{S}))\leq\varepsilon.$
\label{thm:concentration}
\end{theorem}
\vspace{-.1in}
\myparagraph{Remark.} With the convergence theorem, it is not hard to show the following statement: Given sufficiently many samples, if we observe a gap in topology (persistent homology) between the estimated Trojaned and clean models, the gap likely also exists between the true models.

\vspace{-1mm}
\section{Application: A Topological Trojan Detector in Data-Limited Setting}
\label{sec:detector}
\vspace{-2mm}
In this section, we introduce an automatic Trojan detection algorithm based on our observation about Trojaned models' topological abnormality. The Trojan detection problem is essentially a classification problem. Given a set of training models, each clearly tagged as Trojaned or not, can we learn a classifier to predict whether a test model is Trojaned or not. Based on our previous study, we believe topological features can differentiate Trojaned models from clean ones. Our idea is to extract topological features from these models, and use them to train a classifier to predict the Trojan status of a test model. We have in total 12 topological features, including maximum persistence and average death (see Appendix \ref{supp:sec:cycle} for a complete list ). We use a standard MLP (multi-linear perceptron) classifier. 

The major challenge is the limitation of data access. In practice, the Trojaned dataset will not be available to users. We adopt the data-limited setting: for each model (training or testing), only a few clean sample images are given.
To acquire sufficient samples to estimate the correlation of each model, 
we apply a pixel-wise perturbation strategy. A formal algorithm of this is provide in the supplemental material (Appendix \ref{supp:sec:exp}). Given a clean sample image, we iterate through every pixel (or a small patch) modify its value. Then such modified examples are all provided to the model as samples for building the correlation matrix. 

\begin{figure}[htb]
\begin{minipage}{0.48\textwidth}
\centering
\includegraphics[scale=0.31]{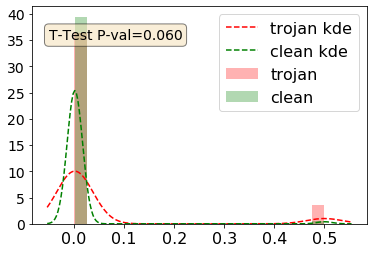}\\
(a). Ave DeathTime - Perturbed Inputs
\end{minipage}
\begin{minipage}{0.48\textwidth}
\centering
\includegraphics[scale=0.31]{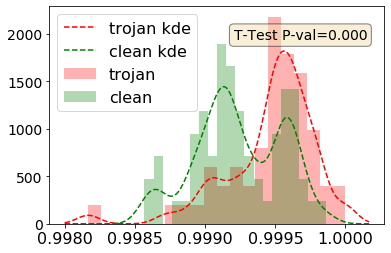}\\
(b). Max-Persist - Perturbed Inputs
\end{minipage}
\caption{Ideal Feature Distribution v.s. Practical Feature Distribution. (a). Average death time calculated using real Trojaned data. (b). Average death time calculated using pixel-wise perturbed data. (c). Maximum persistence calculated using real Trojaned data. (d). Maximum persistence calculated using perturbed data.}
\label{fig:implement}
\end{figure}

To confirm that this sampling strategy is sufficient in mining the topological structure, we carry out the same hypothesis testing as in Sec.~\ref{sec:empirical-study}, except that we use the perturbed samples instead of the Trojaned dataset. As shown in Fig.~\ref{fig:implement}, we still observe significant topological difference between the Trojaned and clean models. This gives us sufficient confidence to use topological features for Trojan detection, with the perturbed samples.

Formally, we propose our Trojaned network detection algorithm in Alg.~\ref{alg:code}. 
We validate our Trojan detector on synthetic and competition datasets, comparing with SoTA baselines.

\begin{algorithm}[!h] 
	\caption{Topological Abnormality Trojan Detection}
	\label{alg:code}
	\begin{algorithmic}[1]
		\State {\bfseries Input:} Training set models $\{f_1, f_2, \cdots, f_N\}$, Testing input associated with each model $X=\{X_1, X_2, \cdots, X_N\}$, Ground truth indicating Trojaned or not $Y=\{y_1, y_2, \cdots, y_N\}$
		\State {\bfseries Output:} Trojaned model detector $g$
		\For{$i=1,\cdots,N$}
        \State {$X_i'$ = Pixel-wise Perturb $(X)$}
		\State Calculate correlation matrix $M(f_i, X_i')$
		\State Build filtration of VR complex $\emptyset \subseteq \mathcal{S}_{t_1 } \subseteq \mathcal{S}_{t_2} \subseteq \mathcal{S}_{t_T} $ using $M(f_i, X_i', \rho)$
		\State Extract topological feature $z_i(\mathcal{S})$ 
		as described in section \ref{paragraph:topo_feature}
		\EndFor
	\State Train Trojan detector $g$ with features $Z=\{\vz_1, \vz_2, \cdots, \vz_{f_N}\}$ and Label $Y$
	\end{algorithmic}
\end{algorithm}

\myparagraph{Synthetic Dataset Experiment.} We generate our synthetic dataset using NIST trojai toolkit\footnote{https://github.com/trojai/trojai}. In synthetic datasets, we trained 140 LeNet5 \cite{lecun1998lenet5} and 120 ResNet18 \cite{he2016resnet} with MNIST \cite{lecun1998lenet5} separately. We also trained 120 ResNet18 and 120 Densenet121 \cite{huang2017densenet} with CIFAR10 \cite{krizhevsky2009cifar} separately. Half of these models are trained with Trojaned datasets where we manually applied $20\%$ one-to-one Trojan attack. Specifically, for Trojaned databases, we picked one of the source class and add a reverse-lambda shape trigger (Figure \ref{fig:Trojan_attack}) to a random corner of the input images. Then we changed the edited images' class to a predetermined target class and mixed them into the training database. Trojaned models are trained with these pollutant database and clean models are trained with original clean database. Furthermore, Trojaned models trained with MNIST datasets are constrained to maintain at least $95\%$ successful attack rate (frequency of predicting the target class when trigger is presented on the test image) and models trained with CIFAR10 are constrained to maintain at least $87\%$ successful attack rate. At the same time, MNIST models and CIFAR10 models also need to maintain at least $97\%$ and $80\%$ testing accuracy on clean inputs separately. There are no significant difference in terms of testing accuracy between clean models (on average $99\%$ for MNIST and $84\%$ for CIFAR10) and Trojaned models (on average $99\%$ for MNIST and $84\%$ for CIFAR10). \label{paragraph:data_generation} 

We compare our Trojan detector's performance with several commonly cited approaches. (1) Neural cleanse (NC) \cite{wang2019cleanse}, (2) Data-limited Trojaned network detection (DFTND) \cite{wang2020DFTND}, (3) Universal litmus pattern (ULP) \cite{kolouri2020ulp}. (4) Baseline classifier using correlation maxtrix directly (Corr). We evaluate using AUC (area under the curve) and ACC (accuracy). More experimental details are provided in supplemental material (Appendix \ref{supp:sec:exp}). Table \ref{table:sysnthetic_experiment} shows the results. We observe consistently that our method is superior compared with other baselines. Making highly accurate prediction of Trojan status of test models. Our method outperforming baseline (4) shows that the short cut phenomenon cannot be directly capture by inspecting the correlation matrix. But our topological approach can capture it.

\begin{table*}[thb]
\caption{Detection Performance on Synthetic Datasets}
\begin{center}
    \resizebox{1.0\columnwidth}{!}{
        \begin{tabular}{l|c|cccc|c}
            \hline
            Dataset & Criterion & NC & DFTND & ULP & Corr & Topo\\
            \hline
            \multirow{2}{*}{MNIST+LeNet5} & ACC & $0.50\pm0.04$ & $0.55\pm0.04$ & $0.58\pm0.11$ & $0.59\pm0.10$ & $\bf{0.85\pm0.07}$\\
            & AUC & $0.48\pm0.03$ & $0.50\pm0.00$ & $0.54\pm0.12$ & $0.62\pm0.10$ & $\bf{0.89\pm0.04}$\\
            \multirow{2}{*}{MNIST+Resnet18} & ACC & $0.65\pm0.07$ & $0.53\pm0.07$ & $0.71\pm0.14$ & $0.56\pm0.08$ &  $\bf{0.87\pm0.09}$\\
            & AUC & $0.64\pm0.11$ & $0.50\pm0.00$ & $0.71\pm0.14$ & $0.55\pm0.08$ & $\bf{0.97\pm0.02}$\\
            \hline
            \multirow{2}{*}{CIFAR10+Resnet18} & ACC & $0.64\pm0.05$ & $0.51\pm0.10$ & $0.56\pm0.08$ & $0.72\pm0.07$ & $\bf{0.93\pm0.06}$\\
            & AUC & $0.63\pm0.06$ & $0.52\pm0.04$ & $0.55\pm0.05$ & $0.81\pm0.08$ & $\bf{0.97\pm0.02}$\\
            \multirow{2}{*}{CIFAR10+Densenet121} & ACC & $0.47\pm0.02$ & $0.59\pm0.07$ & $0.55\pm0.12$ & $0.58\pm0.07$ &  $\bf{0.84\pm0.04}$\\
            & AUC & $0.58\pm0.12$ & $0.60\pm0.09$ & $0.52\pm0.02$ & $0.66\pm0.07$ & $\bf{0.93\pm0.03}$\\
            \hline
        \end{tabular}
        \label{table:sysnthetic_experiment}
    }
\end{center}
\vspace{-3mm}
\end{table*}

\myparagraph{Competition Dataset Experiment.}
We also test our methods using IARPA/NIST trojai competition public dataset \cite{siderius2018nist}\footnote{https://pages.nist.gov/trojai/docs/data.html}. These datasets consist of synthetic traffic sign images superimposed on road background images. There are 3 architectures (ResNet50, DenseNet121, InceptionV3) in Round1 data. Here we show our method's performance using ResNet and DenseNet only. In this dataset, a randomly generated polygon shape Trojan trigger (Figure \ref{fig:Trojan_attack}-(a)) is overlayed on top of the foreground of $5\% \sim 50\%$ of training examples. The Trojaned model will predict the target class whenever a trigger is presented on the images for classes (all-to-one attack). All models have fixed 5 classes output. There are around 200 clean input images are given as reference for each of these models. 

For competition dataset, we leave NC run with its default setting. To finish the experiment in a reasonable amount of time, we randomly pick 200 models from training to search for the optimal threshold for DFTND. For ULP, instead of looping through all models in every epoch, we randomly sampled a batch of 500 models for training. Following table shows the performance. Our method performs superior in this dataset.

\begin{table*}[thb]
\caption{Detection Results on Synthetic Datasets}
\vspace{-3mm}
\begin{center}
    \resizebox{1.0\columnwidth}{!}{
        \begin{tabular}{l|cccc|c}
            \hline
            Dataset & Criterion & NC & DFTND & ULP & Topo\\
            \hline
            \multirow{2}{*}{Round1-ResNet} & ACC & $0.63\pm0.03$ & $0.38\pm0.05$ & $0.63\pm0.00$ & $\mathbf{0.77\pm0.04}$\\
            & AUC &  $0.56\pm0.01$ & $0.45\pm0.05$ & $0.62\pm0.03$ & $\mathbf{0.87\pm0.03}$\\
            \multirow{2}{*}{Round1-DenseNet} & ACC & $0.47\pm0.05$ & $0.49\pm0.04$ & $\mathbf{0.63\pm0.06}$ & $0.62\pm0.04$\\
            & AUC & $0.42\pm0.03$ & $0.51\pm0.01$ & $0.63\pm0.06$ & $\mathbf{0.69\pm0.04}$\\
            \hline
        \end{tabular}
    }
\end{center}
\vspace{-6mm}
\end{table*}

\vspace{-0.1mm}
\section{Conclusion}
\vspace{-1mm}
In this paper, we inspected the structure of Trojaned neural networks through a topological lens. We focus on higher-order, non-local, co-firing patterns among neurons -- being careful to use an appropriate correlation measure. In particular, we observed -- and statistically verified -- the existence of robust topological structures  differentiating between the Trojaned and clean networks. This revealed an interesting short-cut between shallow and deep layers of a Trojaned model. These topological methodology lead to a development of a highly-competitive method of detecting Trojan attacks. 

More broadly, it appears this method could be adapted to other neural network structure analysis tasks -- and perhaps promises ways of excising the undesirable structures.

\bibliographystyle{plain}
\bibliography{ms.bib}

\appendix

\section{Supplementary Material - Summary}

In this supplemental material, we provide additional details on the theory, the algorithms, and the experiments. In Section \ref{supp:sec:ph}, we provide a formal description of persistent homology, as well as the bottleneck distance. In Section \ref{supp:sec:cycle}, we provide details on how we compute the cycle representatives. In Section \ref{supp:sec:shortcut}, we continue analyzing the population level difference between Trojaned and clean models, with a focus on the short-cuts.  Section \ref{supp:sec:proof} includes the proof of the two theorems on (1) the existence of a topological discrepancy between Trojaned and clean models; (2) the convergence of the estimation of persistent homology in terms of the bottleneck distance. In Section \ref{supp:sec:exp}, we provide more technical details on the experiments, including our sampling method based on pixel-wise perturbation, used baselines, and experiment configuration.

\section{Persistent Homology and Bottleneck Distance}
\label{supp:sec:ph}

In the language of algebraic topology \cite{munkres1984elements,edelsbrunner2010computational}, we can formulate a $p$-chain as a set of $p$-simplices. A boundary operator on a $p$-simplex takes  all its adjacent $(p-1)$-simplices. In particular, the boundary of an edges consists of its adjacent nodes; the boundary of a triangle consists of its three edges. More generally, the boundary of a $p$-chain is the formal sum\footnote{We remark that we focus on homology over the $\Z_2$ field, which is the simplest, but practical, setup. In this case the sums simply correspond to subsets of chains.} of the boundary of all its elements, $\partial(c) = \sum_{\sigma\in c} \partial(c)$. Assume we have $m_p$ $p$-simplices for $p=0,1,2$. If we fix an index of all the $p$-simplices, a $p$-chain uniquely corresponds to a $m_p$-dimensional binary vector. In dimension $p$, the boundary operator can be viewed as an $m_{p-1}\times m_p$ matrix, called the $p$-dimensional \emph{boundary matrix}. It consists of the boundaries of all $p$-simplices, $\partial_p = [\partial(\sigma_1),\cdots,\partial(\sigma_{m_p})]$. It is often convenient to consider one big boundary matrix, whose blocks are the $p$-dimensional boundary matrices.  

\begin{figure}[h]
  \centering
    \includegraphics[width=0.3\textwidth]{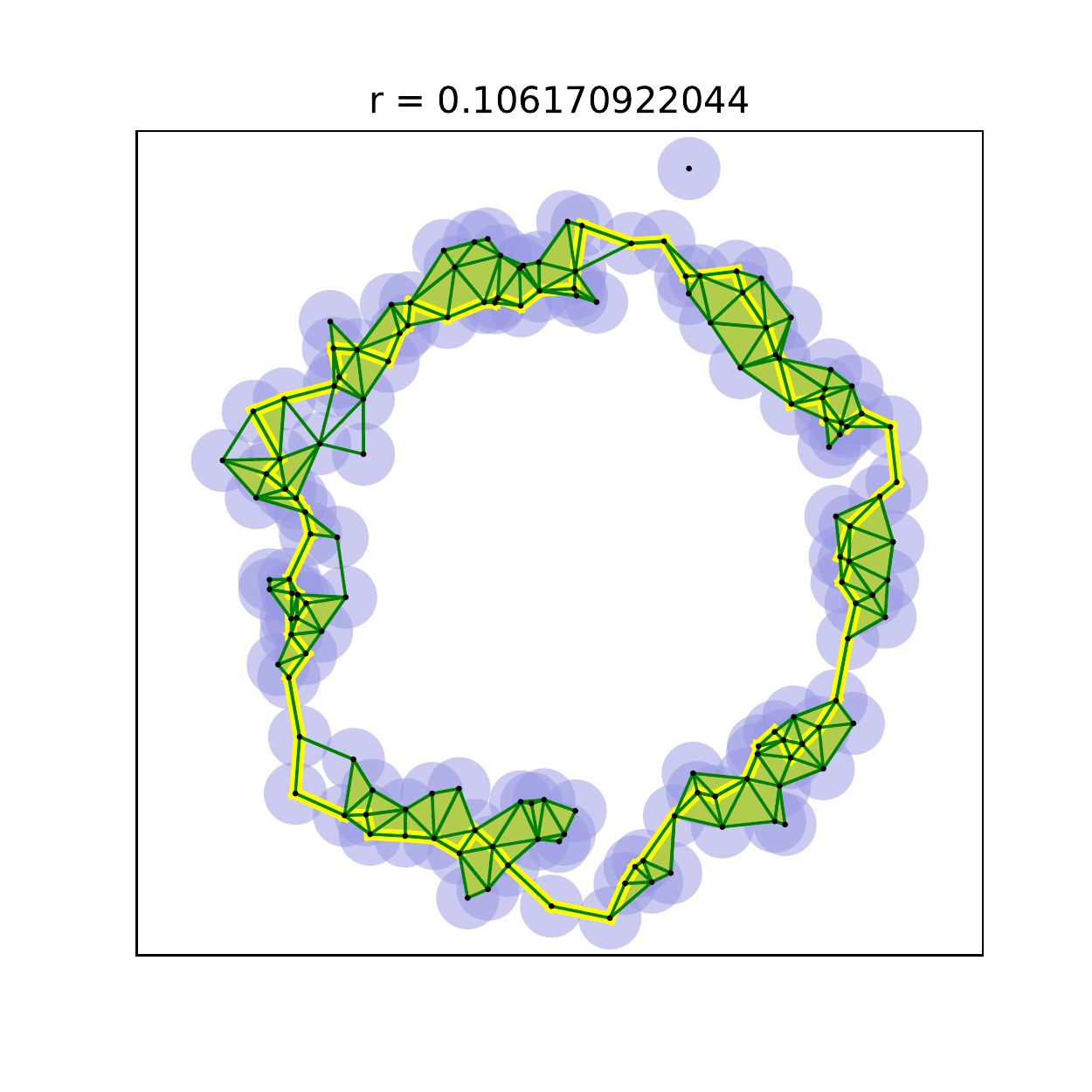} 
    \includegraphics[width=0.3\textwidth]{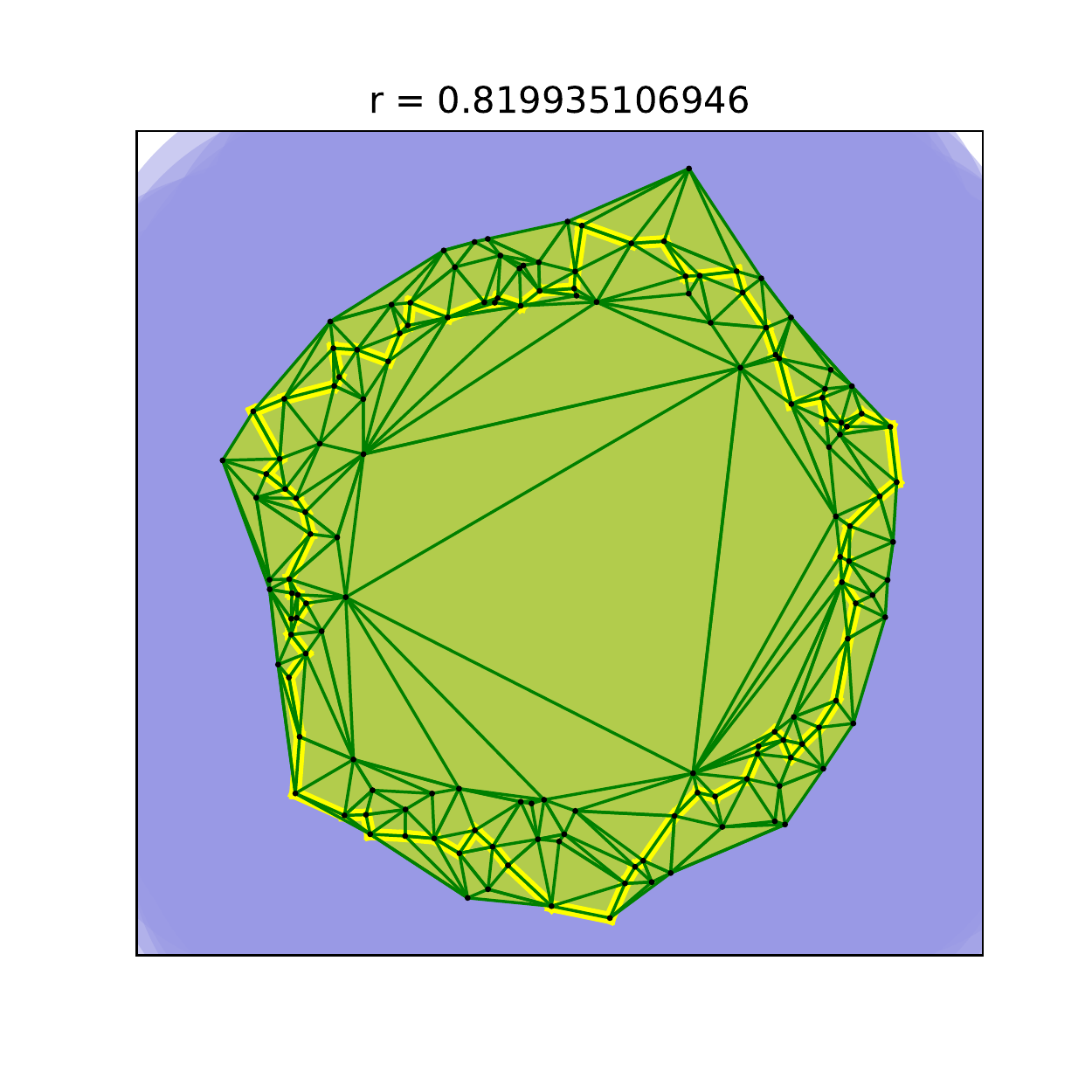}
    \includegraphics[width=0.38\textwidth]{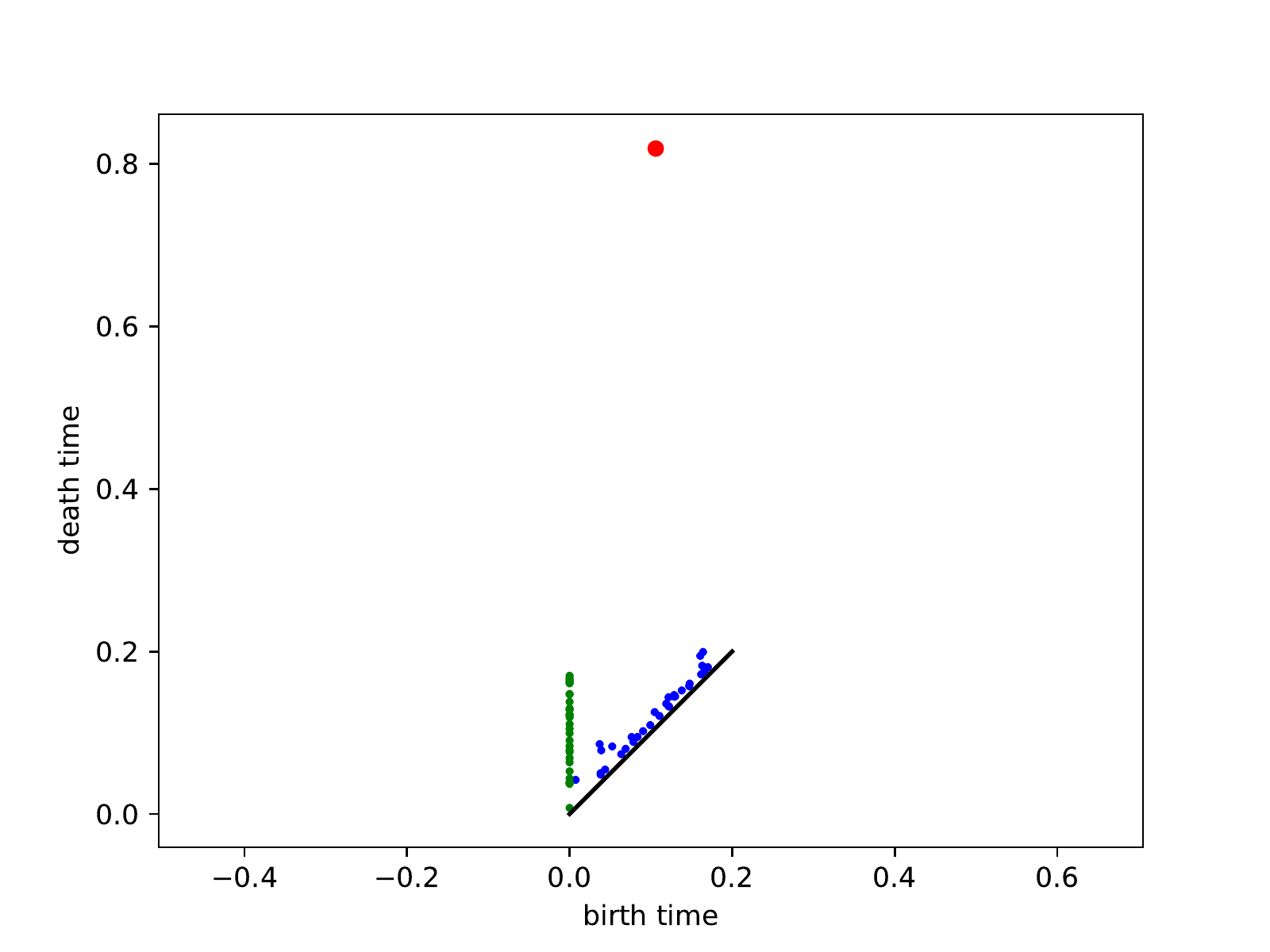}
  \caption{A finite set of points in $\mathbb{R}^2$ sampled with noise from an annulus.
    We see the union of Euclidean balls and the superimposed complex. Its vertices, edges and triangles depict the centers of the balls, pair-wise, and triple-wise intersections at two
    different radii. In our method we use the Vietoris--Rips complex, and here we only show its subcomplex to avoid visual clutter.
    The big loop indicated by the yellow closed curve is born
      on the \emph{left} and dies on the \emph{middle}.
      On the \emph{right} we see the corresponding persistence diagram. The single dot in the upper part corresponds to the prominent feature, namely the big 1-dimensional cycle.
     }
  \label{fig:pers}
\end{figure}

To compute persistent homology, we sort the rows and columns of the big boundary matrix according to the filter function values of the simplices. Then we apply a matrix reduction algorithm, similar to a Gaussian elimination -- except we only allow left-to-right column additions. The classic algorithm \cite{edelsbrunner2010computational} reduces the matrix from left to right, proceeding column by column. After the reduction, the pivoting entries of the reduced matrix correspond to pairs of simplices. We can interpret them as critical simplices that create and kill each topological structure. Their filter function values are the birth and death times of the corresponding persistence dot. The algorithm has worst-case cubic complexity, but modern implementations exhibits linear behaviour on practical inputs. This is an area of active research, and various algorithms were proposed to improve the algorithm either in theory \cite{milosavljevic2011zigzag,chen2013output} or in practice \cite{chen2011persistent, bauer2016phat, bauer2019ripser}. In Figure \ref{fig:pers}, we show sample filtration complexes and the corresponding persistence diagram.

Aside from the birth-death pairs, we also add the set of all points on the diagonal line to the persistence diagram, i.e., $\dgm(M,\calS) = \{(\birth_i,\death_i)\} \cup \{(\birth,\death) | \birth = \death\}$.

\myparagraph{Bottleneck distance.}
We will use the bottleneck distance between two persistence diagrams \cite{cohen2007stability}.
Let $X$ and $Y$ be multisets of points corresponding to two diagrams we plan to compare. Let \mbox{$\Gamma = \{\gamma : X \rightarrow Y\}$} be the family of bijections from $X$ to $Y$. The bottleneck distance is: 

$$
d_b(X, Y) = \inf_{\gamma\in \Gamma}\sup_{\vx \in X} ||\vx - \gamma(\vx)||_{\infty}.
$$
It was shown that the bottleneck distance between diagrams is stable with regard to $L_\infty$ perturbation of the input filter function. Later on, Cohen-Steiner et al.~\cite{cohen2010lipschitz} introduced the $p$-Wasserstein distance between diagrams and showed its stability, when assuming a Lipschitz condition of the filter function.

\section{Cycle Computation}
\label{supp:sec:cycle}
We outline the computations of representative cycles for Vietoris--Rips filtrations. Such cycles play a role in our analysis and interpretation of Trojaned networks. The main goal is to show that these cycle representatives can be computed in a reasonably efficient way for inputs of practical size.

We focus on a particular optimization, in which we extract (homological) cycles from cohomological computations. For computing persistence diagrams (and not cycles) cohomological computations are known to be faster for this type of filtrations \cite{bauer2016phat}. This leads to a simple, efficient technique for extracting cycles, which to the best of our knowledge is novel. We benchmark our implementation and show that it performs well on a variety of practical inputs.

We begin by stating that there are many candidate cycles that can serve as a representative. Finding an ideal representative in terms of certain measurement (e.g., length, area) is a challenging research problem itself \cite{wu2017optimal,zhang2019heuristic,dey2020computing}. We focus on one  way of extracting cycles, worked well in other types of applications, e.g. \cite{wang2021topotxr}. This method is based on certain properties of boundary matrices, which we mention next.

Recall that the classic persistent homology computation reduces the ordered boundary matrix in a column-by-column fashion. In the reduced matrix, each non-zero column corresponds to one persistence dot; the non-zero entries of the column happens to be a cycle representative of this dot. These are simple, well know facts, but not entirely obvious on first glance; see \cite{edelsbrunner2010computational} for an explanation.

These facts allow us to extract the cycles representing any subset of dots we select. One downside is that computing cycles is incompatible with code optimizations used in modern implementations. Indeed, most efficient off-the-shelf software packages do not support cycle extraction. Most notably, Ripser \cite{bauer2019ripser} uses an implicit matrix representation, which makes cycle extraction a non-trivial task. We envision that as the need arises, better cycle-extraction approach will be developed.

\myparagraph{Implementation details.}
Next, we mention some details about our implementation -- aimed at readers familiar with implementational aspects of persistent homology. We focus on the aspect which was not obvious to the authors, namely the strange marriage of cohomological computations and extraction of homological cycles.

For this paper we propose a matrix reduction strategy. This is our way of circumventing the current lack of support for cycle extraction in off-the-shelf software. We use a home-brewed implementation of matrix reduction in tandem with Ripser, which provides the maximum death time, which in turn we use to prune the input for the cycle extraction phase. More explicitly, if the maximum death time is $\epsilon$, we can safely remove all edges with weights greater than $\epsilon$. In the next phase our custom implementation allows for efficient cycle extraction. We give more details of this part next.

We implement a version of the matrix reduction algorithm, working on an explicitly represented (co)boundary matrices. Our implementation is inspired by the PHAT package \cite{bauer2016phat}, uses crucial optimizations introduced therein, but is generally simpler.

In particular, use the bittree data-structure to store and update the reduced column during the reduction. We also switch to cohomology computations in conjunction with the 'twist' optimization. In short, this ensures that for practical inputs the complexity is roughly linear in the number of columns of the matrix -- which is not the case for homological computations; this computation yields exactly the same persistence diagrams. All of these optimizations are known, and summarized in \cite{bauer2016phat}. The one thing we add is the extraction of (homological) cycles, using the information contained in the reduced coboundary matrix (which would naturally yield cocycles).

In broad strokes the new part works as follows: we reduce the coboundary matrix; we then use the information contained in this reduced matrix to generate a pruned boundary matrix; we then reduce this boundary matrix, and extract the cycles as usual. These cycles are identical to those extracted from the regular boundary matrix. Importantly, the pruned boundary matrix is expected to be significantly sparser, the computations are expected to be significantly faster -- at least in the case of filtrations arising from skeleta of Vietoris--Rips complexes.

We verify the above assumptions by benchmarking our implementation using a selection of practical inputs coming from the problem described in the paper.

\myparagraph{Benchmarks.}
We first stress that an attempt to directly reduce the boundary matrix was generally futile for this kind of data. We hypothesise that the computational complexity may be roughly quadratic in the number of columns (simplices) -- similar behaviour was described in \cite{bauer2016phat}.

We report timings for the two matrix reduction steps present in our method: (1) reduction of the coboundary matrix and (2) the reduction of the sparsified boundary matrix, which yields the cycles. We remark that the columns of the final reduced boundary matrix directly contain the cycles, so no non-trivial cost is involved in extracting them.

Our implementation is written in C++, compiled with g++ version 7.4.0. The experiments were performed on a single core of an Intel(R) Xeon(R) CPU E5-2650 v4 @ 2.20GHz using 250GB RAM. Each example used 1496 input points (i.e., a correlation matrix of 1496 neurons). Table \ref{table:bench} shows measurements for a subset of models used in the paper.

\begin{table}[htb]
\caption{The table presents timings and other statistics related to our method.
Column "cutoff" shows the parameter controlling the maximal edge weight, which is specific for a particular dataset;
"cobd red." and "bd red." show the timings for the coboundary and boundary matrix reduction times, in seconds;
"nonzero" shows the number of nonzero elements in the sparsified boundary matrix.
The entries of the table are sorted by the number of simplices, from high to low.\\
}
\label{table:bench}
\begin{tabular}{llllllll}
input-id & cutoff    & num simplices  & cobd red (s) & bd red (s) & nonzero \\
\hline
id-200 & 1.0001 & 558013236 & 116.038 & 4.62077 & 3353285 \\
id-217 & 1.0001 & 558013236 & 140.795 & 4.69326 & 3353285 \\
id-223 & 1.0001 & 558013236 & 124.463 & 8.91073 & 3353285 \\
id-282 & 1.0001 & 558013236 & 147.016 & 5.60385 & 3353285 \\
id-213 & 0.530726 & 507214674 & 128.617 & 4.94272 & 3163966 \\
id-299 & 0.64059 & 390774143 & 102.914 & 3.52663 & 2800161 \\
id-285 & 0.474288 & 244317051 & 55.3774 & 2.89892 & 2112814 \\
id-292 & 0.613246 & 169420317 & 34.8003 & 1.78646 & 1744824 \\
id-251 & 0.371385 & 136133374 & 24.5781 & 1.2683 & 1469012 \\
id-235 & 0.468028 & 108670836 & 26.6361 & 1.09387 & 1375302 \\
id-226 & 0.4699 & 83133043 & 13.2957 & 0.822209 & 1212238 \\
id-306 & 0.586819 & 19978601 & 7.48968 & 1.01139 & 1790934 \\
id-215 & 0.523341 & 12693165 & 15.3573 & 2.7157 & 2678062 \\
id-287 & 0.677098 & 10734370 & 6.20011 & 2.32303 & 2635066 \\
id-252 & 0.670271 & 3071377 & 5.54643 & 2.24891 & 2498068 \\
id-232 & 0.672349 & 2461912 & 3.72422 & 1.42269 & 1450699 \\
id-259 & 0.706322 & 1459357 & 0.953129 & 0.878887 & 640487 \\
id-286 & 0.660859 & 938347 & 0.361241 & 0.27219 & 4580 \\
id-276 & 0.60397 & 841659 & 0.270636 & 0.316308 & 3016 \\
id-266 & 0.664085 & 664494 & 0.226013 & 0.214201 & 100136 \\
\end{tabular}
\end{table}

We observe the extra reduction step increases the time at most by a factor of 2; however in most cases this extra cost is negligible.
What we do not report is the extra cost related to the creation the sparsified boundary matrix. However the second matrix typically very sparse, contains many empty columns, and its creation is fast; see the rightmost column in the table for the number of nonzero entries in the matrix.

The only real downside of our method in the current implementation is the explicit storage of the (co)boundary matrices. On the other hand, our result gives hope that modern tools like Ripser can be enriched with efficient cycle extraction.

We also mention that the computations are highly data-dependent. Over all datasets the mean of the  reduction total time was 28.05 seconds, with standard deviation of 45.78 seconds. We showed a representative subset of these datasets. 

It is also apparent that some of the sparsified boundary matrices are quite hard to reduce, relative to the number of nonzero elements. In this sense, we were lucky that this behaviour only manifested for very sparse matrices. Overall the properties of our ad-hoc method require further investigation.

In any case, the above shows that the extraction of (homological) cycles can be efficiently performed on real world datasets, avoiding the mentioned pitfall of quadratic running time. The data generated in this way undergoes statistical analysis in the next section.

\section{Statistical Inference Results on the Shortcut}
\label{supp:sec:shortcut}

In this section, we further investigate the existence of shortcuts through statistical analysis. In the main paper, we found that the Trojaned model can be identified through both 0D and 1D persistence diagrams, i.e., average death time of 0D diagram and maximum persistence of 1D diagram. Based on this, we hypothesize edges relevant to these topological features can be the shortcuts.
For 0D, a shortcut could be an edge that kills connected component during the filtration. The filter function value ($1-\rho_{i,j}$) of such edge is the death time. As shown in the main paper, the average death time of 0D diagrams clearly separates Trojaned and clean models. For 1D, a shortcut could also be the longest edge (i.e., the edge crossing the most layers of a neural network) in the high persistence 1D cycles. 

We use persistent homology to select these shortcut candidates, and compare the length of these shortcut candidates from Trojaned models and clean models.
Formally, we measure the \emph{length} of an edge as the number of layers that an edge crossed (the index of layer contains the terminal neuron minus the index of layer contains the beginning neuron). 
For the purpose of verification, we use purely Trojaned examples to excite neurons and calculate the correlation matrix and VR filtration. 

\begin{figure}[!htb]
\begin{minipage}{0.45\linewidth}
    \centering
    \includegraphics[width=\textwidth]{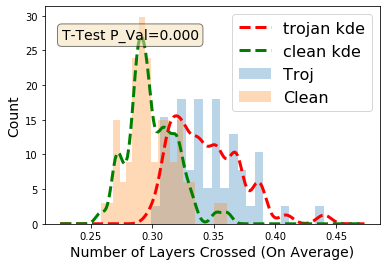}
    (a)
\end{minipage}
\begin{minipage}{0.45\linewidth}
    \centering
    \includegraphics[width=\textwidth]{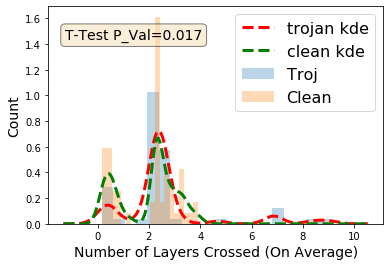}
    (b)
\end{minipage}
\caption{Distribution of average lengths of shortcut edge candidates. (a). Average length of death edges in 0D persistence. Trojaned models generally have longer death edges. (b). Average length of the longest edges in high persistent 1D cycles. In Trojaned models, these edges are longer than in clean models.}
\label{fig:more_hypothesis}
\end{figure}

We first find all edges that kill a 0D homology class (called death edges) and measure their average length. The distribution of the average death edge length is displayed in Figure \ref{fig:more_hypothesis}-(a). 
For each model, we only use the top 1000 edges (w.r.t.~death time). 
Note many edges are connecting neurons within a same layer and have 0 length. As a result, their average length can be smaller than 1. 
We observe a significant difference between average lengths of the death edges of Trojaned and clean models. Trojaned models tend to have longer death edges.
The two sample independent t-test average death edge length between Trojaned models and clean models is rejected at significance level smaller than $\leq 0.001$. Due to the computation precision reason, we round the p-value to the 4th digit. In reality, the significance level should be smaller than 0.001.

For 1D homology, we collect the longest edge in high persistence 1D homology cycles, computed from the algorithm presented in Section \ref{supp:sec:cycle}. For each model, we collect top 500 persistent 1D cycles. We extract the longest edges of these cycles and take their average length.  The distribution of the average length of these edges for different models is presented in Figure \ref{fig:more_hypothesis}-(b). Similar to in 0D, the average length is generally low as we are averaging over many cycles. The distribution for clean models presents a bi-modal shape while the Trojaned models' is right skewed. On average, these type of edges in Trojaned models bypass more layers than those in clean models. The t-test's result corroborate our conclusion (p-value = 0.017), meaning a significance level of 95\%.

\begin{figure}[!htb]
    \centering
\begin{tabular}{ccc}
    \includegraphics[width=.3\textwidth]{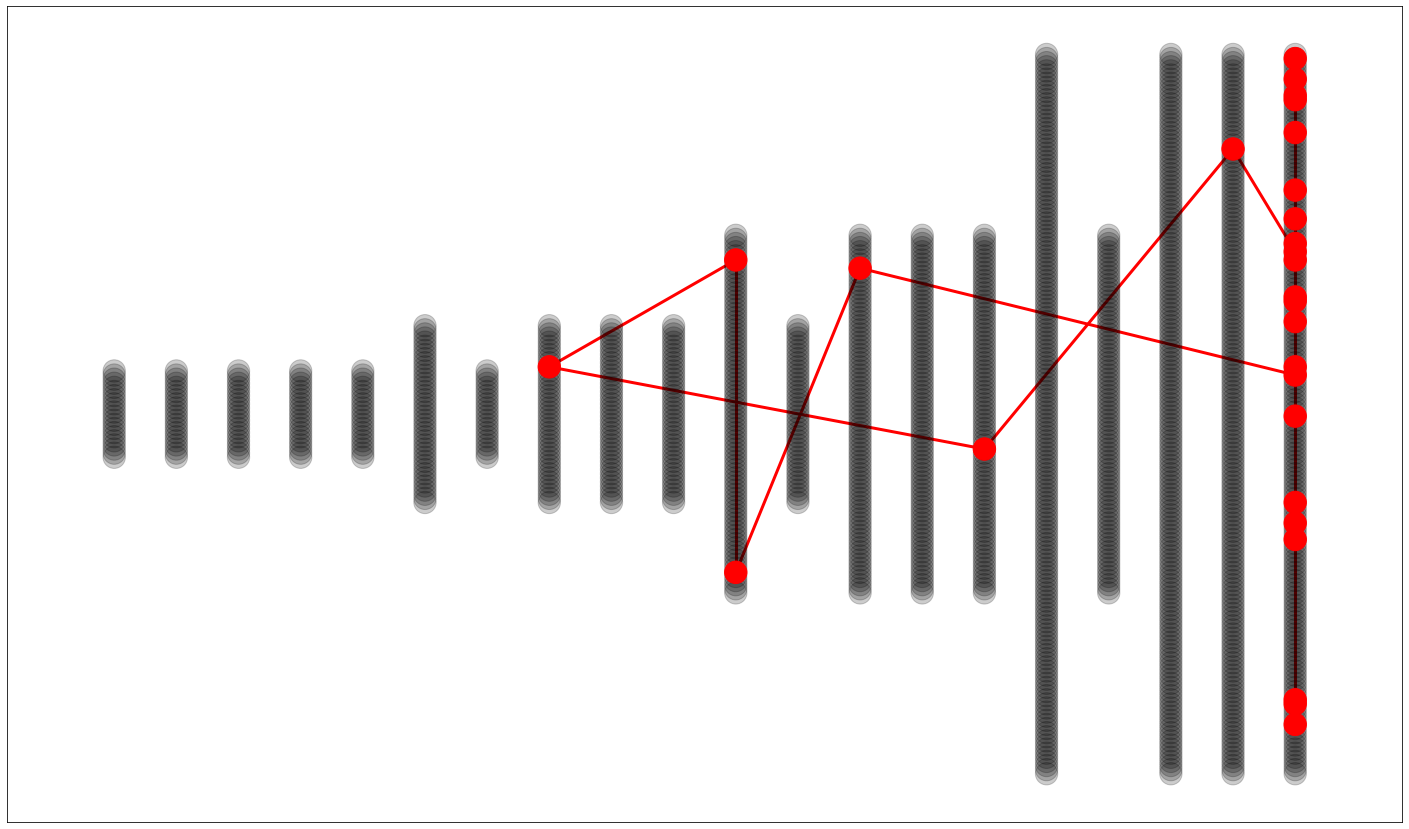} &
    \includegraphics[width=.3\textwidth]{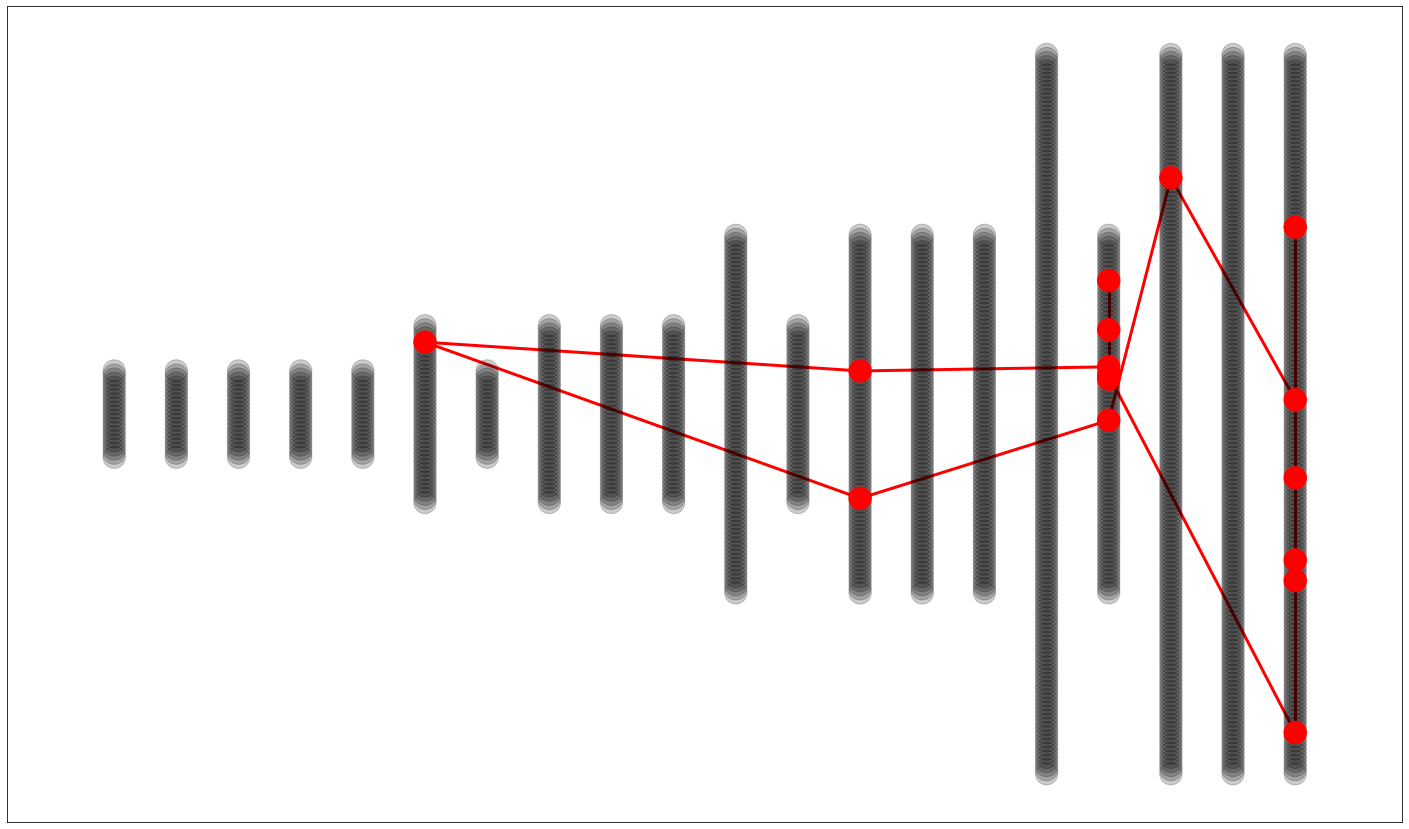} &
    \includegraphics[width=.3\textwidth]{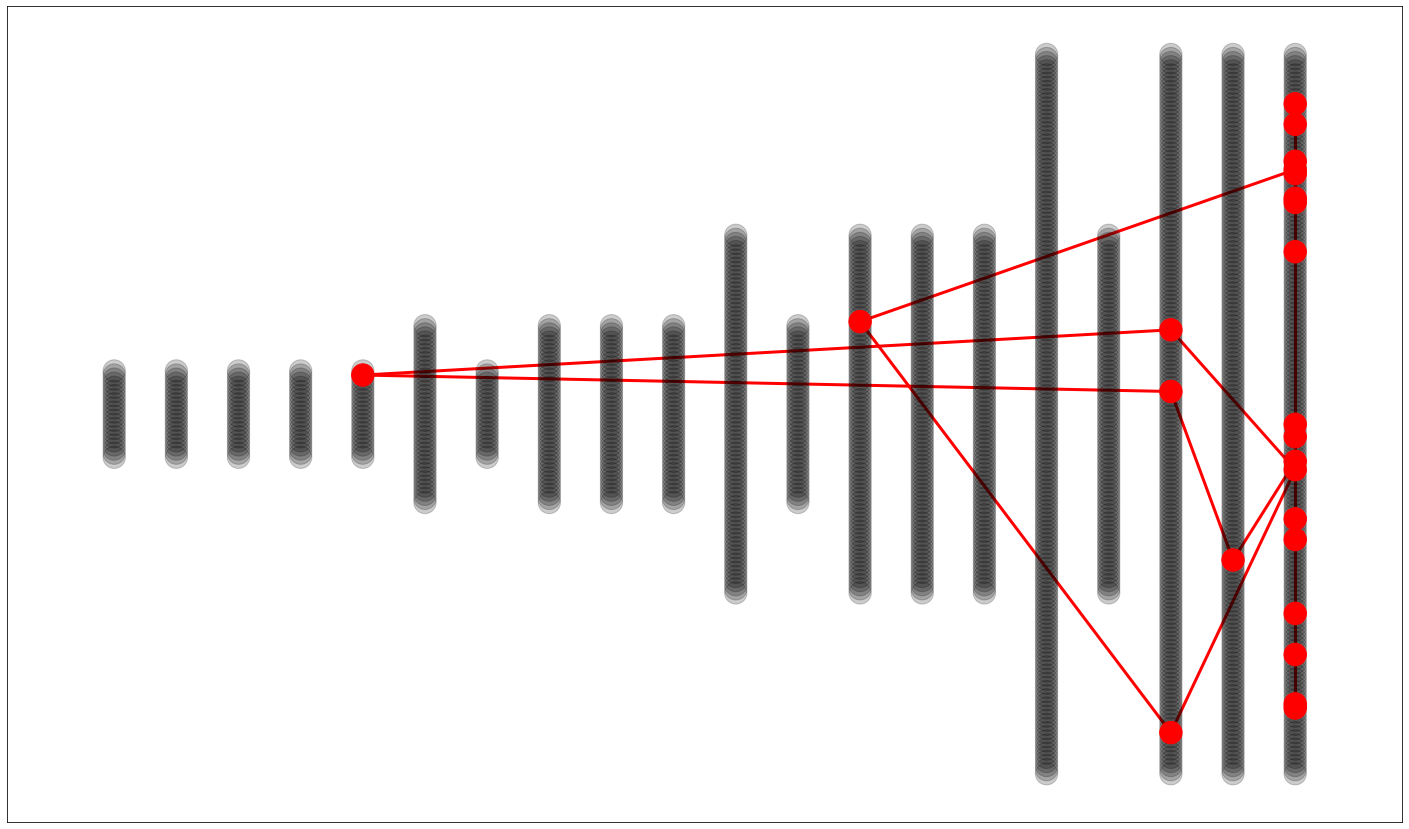}
\end{tabular}
\caption{More examples of the top persistence cycles from Trojaned models.}
\label{fig:more-cycles}
\end{figure}

We note that even though not all Trojaned models have significantly longer short cut edges than clean models, we do discover a significant subset of Trojaned models (about 10 out of 70) with long short cut edges from the top persistence cycles. We show a few samples in Figure \ref{fig:more-cycles}. The behavior of these Trojaned models and their difference with the rest is yet to be further studied.

\section{Theorems and Proofs}
\label{supp:sec:proof}
In this section, we provide proof of the two theorems in the main paper. 

\subsection{Proof: Existence of Topological Discrepancy}
\begin{figure}[H]
\centering
\begin{tabular}{c c c}
\centering
     \includegraphics[scale=0.15]{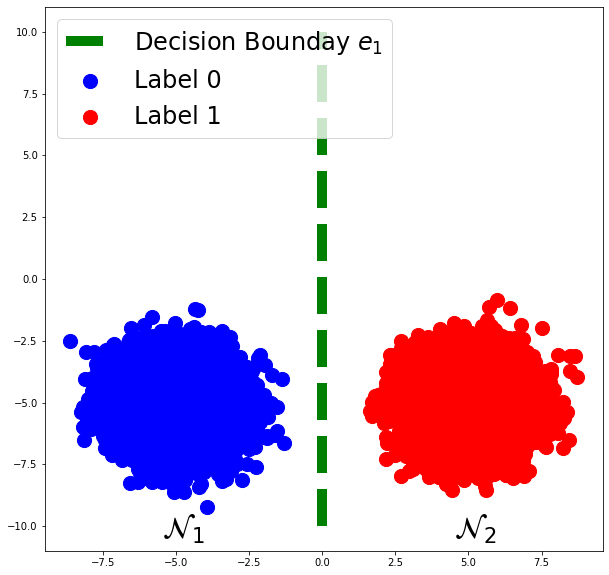}&
     \includegraphics[scale=0.15]{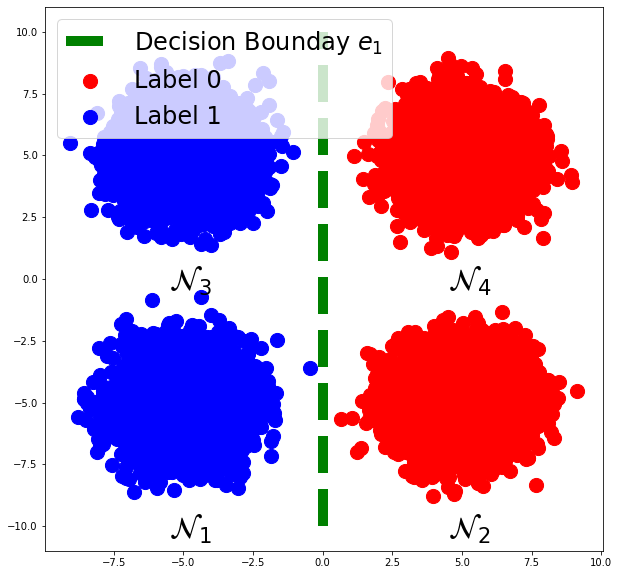}&  
     \includegraphics[scale=0.15]{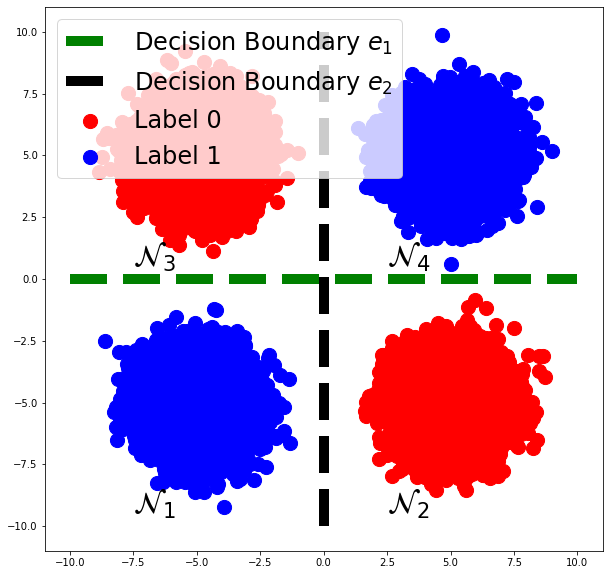}\\
     (a)  $\mathcal{D}_1$ (Original Data) &
     (b)$\mathcal{D}_2$ (Trojaned Feature with Clean Labels) &  
     (c) $\mathcal{D}_3$ (Trojed Data)
\end{tabular}
\caption{Demonstration of the Trojaned Gaussian pair. (a). $\mathcal{D}_1$ is the original data distribution. (b). $\mathcal{D}_2$ is the mixture data distribution of original distribution and the shifted distribution caused by trigger overlaying (note the classification risk is not necessary to be 0 to separate any two of these 4 Gaussian distribution). (c).$\mathcal{D}_3$ is the Trojaned dataset where labels will be modified for those Trojaned examples.}
\label{fig:proof_gaussian_examples}
\end{figure}

Recall the definition of Trojaned Mix-Gaussian Pair. An illustration can be found in Figure~\ref{fig:proof_gaussian_examples}

\begin{definition}[Trojaned Mix-Gaussian Pair] 
\label{def:mix_pair_2}
Let $\mu_1 = 2(-e_2-e_1)\sigma \sqrt{\log (\frac{1}{\eta})} $, $\mu_2 =  2(-e_2+e_1)\sigma \sqrt{\log (\frac{1}{\eta})} $, $\mu_3 =  2(e_2-e_1)\sigma \sqrt{\log (\frac{1}{\eta}) }$, $\mu_4 =  2(e_2+e_1)\sigma \sqrt{\log (\frac{1}{\eta}) }$. Let $i\sim unif(\{1,2\})$ and $j\sim unif(\{1,2,3,4\})$. We define the following pair of distributions $(\mathcal{D}_1, \mathcal{D}_2,\mathcal{D}_3)$ to be Trojaned Mix-Gaussian Pair (see supplementary section \ref{supp:sec:ph}), where:
\begin{align*}
 &\mathcal{D}_1 \text{(Original data)}
 =\{(\vx, \vy) : x \sim \mathcal{N}(\mu_i, \sigma^2 I_{d}), \hspace{1mm} \vy=\text{i MOD 2}\}\\
 &\mathcal{D}_2 \text{(Trojaned feature with correct labels)}
 =\{(\vx, \vy) : x \sim \mathcal{N}(\mu_i, \sigma^2 I_{d}), \hspace{1mm} \vy=\text{j MOD 2}\}\\
 &\mathcal{D}_3 \text{(Trojaned feature with modified labels)}
 =\{(\vx, \vy) : x \sim \mathcal{N}(\mu_i, \sigma^2 I_{d}), \hspace{1mm} \vy=\mathbbm{1}_{j\in\{2,3\}}\}
\end{align*}
\end{definition}
We study the hypothesis class $\mathcal{H}$ of binary output neural networks with two hidden layers and four neurons in each hidden layer equipped with an indicator activation function.

\begin{theorem} 
Let $(\mathcal{D}_1,\mathcal{D}_2,\mathcal{D}_3)$ be Trojaned Mix-Gaussian Pair and $\mathcal{H}$ be the hypothesis class defined as above. Let $R(f,x,y) = \mathbbm{1}\left({f(x)\neq y}\right)$. There exists $f_1,f_2 \in \mathcal{H}$ where 
$\mathbb{E}_{(x,y)\sim \mathcal{D}_1} [R(f_1)] \leq \eta$, \hspace{1mm}
$\mathbb{E}_{(x,y)\sim \mathcal{D}_3} [R(f_2)] \leq \eta$\hspace{1mm}, 
$\mathbb{E}_{(x,y)\sim \mathcal{D}_2} [R(f_2)] \geq \frac{1}{2}$\hspace{1mm}, such that:
$$d_b[Dg(M(f_1, \mathcal{D}_2), \mathcal{S}) - Dg(M(f_2,  \mathcal{D}_2), \mathcal{S})] \geq 0.9$$
where $d_b$ is bottleneck distance. $Dg(M(f_i, \mathcal{D}_2), \mathcal{S})$ is the 1D persistence diagram of the Vietoris--Rips filtration $\mathcal{S}$ that is built on top of the correlation matrix $M(f_i, \mathcal{D}_j)$.  
\end{theorem}

\textbf{Proof}:
The proof is constructive.
Let $f_1, f_2$ be parametrized by $U_1,V_1,W_1, b_1^U,b_1^V,b_1^W$ and $U_2,V_2,W_2, b_2^U,b_2^V,b_2^W$ and let 

\begin{equation}
\begin{aligned}
&U_1 = \left[ \begin{array}{cc}
      e_1^\top  \\
      -e_1^\top  \\
      e_1^\top   \\
      -e_1^\top  \end{array}\right]
      & V_1 = \left[ \begin{array}{cccc}  1& 0 & 0& 0 \\ 0& 1 & 0& 0 \\ 0&0 &1& 0 \\ 0& 0& 0& 1    \end{array}\right] 
      & W_1 = \left[ \begin{array}{cc}  0& 1  \\  1& 0 \\  0& 1 \\  1& 0    \end{array}\right]  \\ 
      & b_1^U = \left[ \begin{array}{cc} 0  \\ 0  \\ 0  \\ 0 \end{array}\right] 
      & b_1^V = \left[ \begin{array}{cc}  -1 \\  -1  \\ -1  \\ -1  \end{array}\right]
      & b_1^W = \left[ \begin{array}{cc}  0  \\  0  \\ 0  \\ 0  \end{array}\right]
\end{aligned}
\end{equation}

\begin{equation}
\begin{aligned}
&U_2 = \left[ \begin{array}{cc}
      e_1^\top  \\
      -e_1^\top  \\
      e_2^\top   \\
      -e_2^\top  \end{array}\right]
      &V_2 = \left[ \begin{array}{cccc}  1& 0 & 1& 0 \\ 1& 0 & 0& 1 \\ 0& 1 & 1& 0 \\ 0& 1 & 0& 1    \end{array}\right] 
      &W_2 = \left[ \begin{array}{cc}  0& 1  \\  1& 0 \\  0& 1 \\  1& 0    \end{array}\right]  \\ 
      &b_2^U = \left[ \begin{array}{cc} 0  \\ 0  \\ 0  \\ 0 \end{array}\right] 
      &b_2^V = \left[ \begin{array}{cc}  -2 \\ -2  \\ -2  \\ -2  \end{array}\right]
      &b_2^W = \left[ \begin{array}{cc}  0  \\  0  \\ 0  \\ 0  \end{array}\right]
\end{aligned}
\end{equation}
One can see $f_1(x)  = \mathbbm{1}_{x^\top e_1<0}$ and $f_2(x)  =\mathbbm{1}_{x^\top e_1 x^\top e_2 \geq 0}$ are Bayes optimal classifier for $\mathcal{D}_1$ and $\mathcal{D}_3$. Since $\|\mu_i- \mu_j\| \geq 4\sigma \sqrt{\log(\frac{1}{\eta})}$, the Bayes risk is at most $\eta$ which implies    $\mathbb{E}_{(x,y)\sim \mathcal{D}_1} [R(f_1)] \leq \eta$ and  $\mathbb{E}_{(x,y)\sim \mathcal{D}_3} [R(f_2)] \leq \eta $.
If we use $\mathbbm{1}_{x^\top e_1 x^\top e_2 \geq 0}$ as decision boundary for classifying $(x,y)$ generated from $\mathcal{D}_2$, due its symmetricity exactly half of the samples will be misclassified thus $\mathbb{E}_{(x,y)\sim \mathcal{D}_2} [R(f_2)] \geq \frac{1}{2}$. Next we analyze the second moment matrix of neurons.
Let $a_1 = \left[\begin{array}{c}
p_1 \\
q_1 
\end{array} \right]$ and $a_2 = \left[\begin{array}{c}
p_2 \\
q_2 
\end{array} \right]$, we next calculate $\mathbb{E}_{x\sim \mathcal{D}_1} [ a_1 \otimes a_1 ] $ and $\mathbb{E}_{x\sim \mathcal{D}_1} [ a_2 \otimes a_2 ]$. 
\begin{equation}
\begin{aligned}
&\mathbb{E}_{x \sim \mathcal{D}_2} [p_1 \otimes p_1]  =  \left[ \begin{array}{cccc}
      \frac{1}{2} & 0 & \frac{1}{2} & 0\\
      0 &\frac{1}{2}  & 0 & \frac{1}{2}\\
      \frac{1}{2} & 0 & \frac{1}{2} & 0\\
      0 & \frac{1}{2 } & 0 & \frac{1}{2}
      \end{array} \right]
      \mathbb{E}_{x \sim \mathcal{D}_2} [q_1 \otimes q_1]  = \left[ \begin{array}{cccc}
      \frac{1}{2} & 0 & \frac{1}{2} & 0\\
      0 &\frac{1}{2}  & 0 & \frac{1}{2}\\
      \frac{1}{2} & 0 & \frac{1}{2} & 0\\
      0 & \frac{1}{2 } & 0 & \frac{1}{2}
      \end{array} \right]\\
      &\mathbb{E}_{x \sim \mathcal{D}_2} [p_1 \otimes q_1]  = \mathbb{E}_{x \sim \mathcal{D}_2} [(q_1 \otimes p_1)^\top ]= \left[ \begin{array}{cccc}
      \frac{1}{2} & 0 & \frac{1}{2} & 0\\
      0 &\frac{1}{2}  & 0 & \frac{1}{2}\\
      \frac{1}{2} & 0 & \frac{1}{2} & 0\\
      0 & \frac{1}{2 } & 0 & \frac{1}{2}
      \end{array} \right]
\end{aligned}
\end{equation}
And $\mathbb{E}_{x \sim \mathcal{D}_2} [p_1] = [\frac{1}{2},\frac{1}{2},\frac{1}{2},\frac{1}{2}]^\top$, $\mathbb{E}_{x \sim \mathcal{D}_2} [q_1] = [\frac{1}{2},\frac{1}{2},\frac{1}{2},\frac{1}{2}]^\top$
\begin{equation}
\begin{aligned}
&\mathbb{E}_{x \sim \mathcal{D}_2} [p_2 \otimes p_2]  =  \left[ \begin{array}{cccc}
      \frac{1}{2} & 0 & \frac{1}{4} & \frac{1}{4}\\
      0 &\frac{1}{2}  & \frac{1}{4} & \frac{1}{4}\\
      \frac{1}{4} & \frac{1}{4} & \frac{1}{2} & 0\\
      \frac{1}{4} & \frac{1}{4} & 0 & \frac{1}{2} 
      \end{array} \right]
      \mathbb{E}_{x \sim \mathcal{D}_2} [q_2 \otimes q_2]  = \left[ \begin{array}{cccc}
     \frac{1}{4} & 0 &0 & 0\\
      0 &\frac{1}{4}  & 0 & 0\\
      0& 0 & \frac{1}{4} & 0\\
      0 & 0 & 0 & \frac{1}{4}
      \end{array} \right]\\
      &\mathbb{E}_{x \sim \mathcal{D}_2} [p_2 \otimes q_2] = \mathbb{E}_{x \sim \mathcal{D}_2} [(q_2 \otimes p_2)^\top ]  = \left[ \begin{array}{cccc}
      \frac{1}{4} & \frac{1}{4} & 0 & 0\\
      0 & 0   &\frac{1}{4}& \frac{1}{4}\\
      \frac{1}{4} & 0 & \frac{1}{4} & 0\\
      0 & \frac{1}{4} & 0 & \frac{1}{4}
      \end{array} \right]
\end{aligned}
\end{equation}
And $\mathbb{E}_{x \sim \mathcal{D}_2} [p_2] = [\frac{1}{2},\frac{1}{2},\frac{1}{2},\frac{1}{2}]^\top$, $\mathbb{E}_{x \sim \mathcal{D}_2} [q_2] = [\frac{1}{4},\frac{1}{4},\frac{1}{4},\frac{1}{4}]^\top$

A simple calculation completes the proof.
\qed 

\subsection{Proof: Convergence Theorem}

\begin{theorem}
 Let $M(f_k, X_k) \in \mathbb{R}^{m_k\times m_k}$ with $m_k \leq m^*, \forall k \in [N]$ and its entries $M_k^{i,j} = \frac{ \psi(v_i(X_k),v_j(X_k))}{\sqrt{ \psi(v_i(X_k),v_i(X_k)) \psi(v_j(X_k),v_j(X_k))}}$ and the its target value $M^*(f_k, \mathcal{D}_k) \in \mathbb{R}^{m_k\times m_k}$ with its entries ${M^*_k}^{i,j} = \frac{\mathbb{E}_{X_k \sim \mathcal{D}_k} \left[\psi(\vv_i(X_k),\vv_j(X_k)) \right]}{\sqrt{\mathbb{E}_{X_k \sim \mathcal{D}_k} \left[\psi(\vv_i(X_k),\vv_i(X_k))\right] \mathbb{E}_{X_k \sim \mathcal{D}_k} \left[\psi(\vv_j(X_k),\vv_j(X_k))\right]}}$
 as defined in section 3 with $\psi(v_i(X),v_j(X)) = \frac{1}{n} \sum_{\vx_l \in X} \psi(v_i(\vx_l),v_j(\vx_l))  $. Suppose $\forall k \in [N], X_k$ are iid sampled from distribution $\mathcal{D}_k $ and  $|\psi(v_i(\vx),v_j(\vx)) | \leq \mathcal{R}$ for all $\vx\sim \mathcal{D}_k,v_i,v_j$, $0< r\leq \mathbb{E}_{\vx \sim \mathcal{D}_k} \psi(v_i(\vx),v_i(\vx))$ for all $i \in [m_k]$, if we have   $\forall k \in [N]$ 
 $$|X_k|  \geq  \frac{16\mathcal{R}^6\left(\log(N)+2\log(m^*)+\log(\frac{1}{\delta}) \right)}{r^4\varepsilon^2} $$ 
 then with probability at least $1-\delta$, for all $k \in [N]$,
 $d_{b}(\dgm(M(f_k, X_k),\mathcal{S}),\dgm(M(f_k, \mathcal{D}_k),\mathcal{S}))\leq\varepsilon$.
\end{theorem}

\textbf{Proof}:
By Hoeffding inequality for each $\psi(\vv_i(X_),\vv_j(X_i))$, if 
$X_k$ has size $n_k \geq  \frac{16\mathcal{R}^6\left(\log(N)+2\log(m^*)+\log(\frac{1}{\delta}) \right)}{r^4\varepsilon^2}$ we have $|\psi(\vv_i(X_k),\vv_j(X_k)) - \mathbb{E}_{X_k \sim \mathcal{D}_k}[\psi(\vv_i(X_k),\vv_j(X_k))]| \leq \frac{\varepsilon r^2}{4\mathcal{R}^2}$ with probability at least $1-\frac{\delta}{{m^*}^2 N} $.

Next we bound
\begin{equation} \label{eq:correlations}
\left| \frac{\psi(\vv_i(X_k),\vv_j(X_k))}{\sqrt{\psi(\vv_i(X_k),\vv_i(X_k)) \psi(\vv_j(X_k),\vv_j(X_k))}} -  \frac{\mathbb{E} \left[\psi(\vv_i(X_k),\vv_j(X_k)) \right]}{\sqrt{\mathbb{E} \left[\psi(\vv_i(X_k),\vv_i(X_k))\right] \mathbb{E} \left[\psi(\vv_j(X_k),\vv_j(X_k))\right]}}\right| 
\end{equation}

By setting $a_1 =\psi(\vv_i(X_k),\vv_j(X_k)), a_2=\mathbb{E} \left[\psi(\vv_i(X_k),\vv_j(X_k)) \right] , b_1 =\psi(\vv_i(X_k),\vv_i(X_k)) ,b_2=\mathbb{E} \left[\psi(\vv_i(X_k),\vv_i(X_k))\right] , c_1 =\psi(\vv_j(X_k),\vv_j(X_k)) , c_2=\mathbb{E} \left[\psi(\vv_j(X_k),\vv_j(X_k))\right]$, we can observe that $\frac{a_1}{b_1c_1} - \frac{a_2}{b_2c_2} = \frac{a_1b_2c_2 - a_2b_1c_1}{b_1c_1b_2c_2}$. Due to the fact that 
$|a_1-a_2| \leq \frac{\varepsilon r^2}{4\mathcal{R}^2}$,
$|b_1^2-b_2^2| \leq  \frac{\varepsilon r^2}{4\mathcal{R}^2}$,
$|c_1^2-c_2^2| \leq \frac{\varepsilon r^2}{4\mathcal{R}^2}$ 
and 
$a_1 \leq \mathcal{R},a_2\leq \mathcal{R}$, 
$r\leq b_2^2 \leq \mathcal{R}, r\leq c_2 \leq \mathcal{R}$, 
we have $b_1c_1b_2c_2 \geq \frac{r^2}{4}$ and $|a_1b_2c_2 - a_2b_1c_1| \leq 2\varepsilon \mathcal{R}^2 $ which implies that  Equation (\ref{eq:correlations}) is bounded by $\varepsilon$.
By taking a union bound on failure probability for all $m_k^2$ entries in matrix $M_k$ and for all $M_k, k \in [N]$ one will get with probability at least $1-\delta$:
$$\forall k \in [N], \|M(f_k, X_k) -  M(f_k, \mathcal{D}_k)\|_{\infty} \leq \varepsilon $$
By stability theorem of bottleneck distance~\cite{cohen2007stability} with probability at least $1-\delta$ for all $k \in [N]$:
\begin{equation*} 
    \begin{aligned}
    &d_{b}(\dgm(M(f_k, X_k),\mathcal{S}) ,  \dgm(M(f_k, \mathcal{D}_k),\mathcal{S})) 
    &\leq \|M(f_k, X_k)-M(f_k, \mathcal{D}_k)\|_{\infty}
    \leq \varepsilon
    \end{aligned}
\end{equation*}
\qed

\section{Experimental Details}
\label{supp:sec:exp}

\subsection{Pixel-wise Perturbation}
\label{app:cp_algorithm}
For Trojan detection, we are only given a few clean samples for each model. We propose a pixel-wise perturbation algorithm to obtain samples. 
\begin{algorithm}[h!] 
	\caption{Pixel-wise Perturbation}
	\label{supp:alg:code}
	\begin{algorithmic}[1]
	\State {\bfseries Input:} Dataset $X=\{\vx_1, \vx_2, \cdots, \vx_m\}$, Number of trials $n$, Input Range $L=\{(\vl_1, \vu_1), (\vl_2, \vu_2), \cdots, (\vl_m, \vu_m)\}$ 
	\State {\bfseries Output:} Coordinate Perturbed Dataset $X'$
	\State {$X'=\emptyset$}
	\For{$i=1,\cdots,\vm$}
		\State {$X'_i=\emptyset$}
		\While{$j \leq n$}
		    \State{$\vx^c_i = \vx_i$}
		    \State{Sample $k \sim \{1,2,\cdots,d\}$, sample a perturbed value $v \sim [\vl_i, \vu_i]$}
		    \State{Set $k$th coordinate of $\vx^c_i[k]=v$}
		    \State{$X_i'=X_i' \cup \vx^c_i$}
		    \State{$j++$}
		\EndWhile
		\State {$X'=X'\cup X_i'$}
	\EndFor
    \State {\bfseries Return:} $X'$
	\end{algorithmic}
\end{algorithm}

\subsection{Synthetic Experiment Baseline Setting and Experiment Configuration}
\label{app:synthetic_setting}

\textbf{Baseline Setting.} We compare our Trojan detector's performance with several commonly cited approaches. Neural cleanse (NC) introduces a reversed engineering approaches where the algorithm tries to find a pattern when overlaying with input can flip the output of the model. It detects a Trojaned model if the median absolute deviation of any resulting reverse engineered pattern goes beyond 2. Data-limited Trojaned network detection (DFTND) identifies a Trojaned model if the difference between the norm of the penultimate layer's representation of a clean input and a adversarial input goes above certain threshold. 
Universal litmus pattern (ULP) adopts a meta training idea where several randomly initialized examples (ULP) are given to all models. These ULPs are optimized to form representations that can be learned by a Trojan detector to discriminate clean and Trojaned models. We also compare with a baseline classifier that exploits the correlation matrix directly (Corr). We extract the top 5 singular values of the correlation matrix and calculate the Frobinius norm of the matrix after thresholding using $25\%, 50\%, 75\%$ percentile of the matrix separately. We combine these values into a feature vector and train a classifier with these feature. 

\textbf{Experiment Configuration.} We use $80\%$ of the data as the training set and use the rest $20\%$ as the testing set. NC  doesn't need training set so we randomly choose $20\%$ of data to measure the performance. DFTND doesn't require training set either. So we use the training set to search for a optimal threshold that minimize the cross entropy loss on training set. We repeat each experiment 5 times and the results are record in Table 1 and Table 2 in the main text. Our detector's performance is consistently better than all baselines.

\end{document}